\documentclass{article}

\usepackage{microtype}
\usepackage{graphicx}
\usepackage{booktabs} 

\usepackage{hyperref}

\usepackage[preprint]{icml2026}
\usepackage{amsmath}
\usepackage{amssymb}
\usepackage{mathtools}
\usepackage{amsthm}
\usepackage{amsfonts}
\usepackage{graphicx}
\usepackage{multirow}
\usepackage{wrapfig}
\usepackage[RGB]{xcolor}
\usepackage{colortbl}
\usepackage{tabularx}

\hypersetup{
        colorlinks=true,
        linkcolor={red!50!black},
        citecolor={blue!50!black},
        urlcolor={blue!80!black}
    }

\definecolor{OUR}{HTML}{885AF4}
\definecolor{PPGS}{HTML}{56BC6C}
\definecolor{PPGP}{HTML}{FFBC6C}

\colorlet{OURTable}{OUR!30}
\colorlet{PPGSTable}{PPGS!30}
\colorlet{PPGPTable}{PPGP!30}
\definecolor{Gray}{gray}{0.9}

\newcolumntype{g}{>{\columncolor{Gray}}c}
\newcolumntype{x}{>{\columncolor{PPGPTable}}c}
\newcolumntype{y}{>{\columncolor{PPGSTable}}c}
\newcolumntype{z}{>{\columncolor{OURTable}}c}

\usepackage[capitalize,noabbrev]{cleveref}

\icmltitlerunning{A Robust PPG Foundation Model using multimodal physiological supervision}

\begin{document}
\twocolumn[
  \icmltitle{A robust PPG foundation model using multimodal physiological supervision}



  \icmlsetsymbol{equal}{*}
  \icmlsetsymbol{intern}{\textdagger}

  \begin{icmlauthorlist}
    \icmlauthor{Eloy Geenjaar}{georgiatech,trends,intern}
    \icmlauthor{Vince Calhoun}{georgiatech,trends}
    \icmlauthor{Scott Daly}{dolby}
    \icmlauthor{Gouthaman KV}{dolby}
    \icmlauthor{Lie Lu}{dolby}
    \icmlauthor{Trisha Mittal}{dolby}
    \icmlauthor{Daniel P. Darcy}{dolby}
  \end{icmlauthorlist}

  \icmlaffiliation{georgiatech}{Department of Electrical and Computer Engineering, Georgia Institute of Technology, Atlanta, USA}
  \icmlaffiliation{trends}{Tri-Institutional Center for Translational Research in Neuroimaging and Data Science (TReNDS), Georgia State University, Georgia Institute of Technology, Emory University, Atlanta, USA}
  \icmlaffiliation{dolby}{Dolby Laboratories, San Francisco, USA}
  \icmlcorrespondingauthor{Eloy Geenjaar}{egeenjaar@gatech.edu}
  \icmlcorrespondingauthor{Daniel P. Darcy}{dan.darcy@dolby.com}

  \icmlkeywords{Machine Learning, ICML}

  \vskip 0.3in
]


\printAffiliationsAndNotice{%
  \textdagger Work completed during an internship at Dolby Laboratories. Authors are ordered alphabetically by last name, except for the corresponding authors. 
}

\begin{abstract}
Photoplethysmography (PPG), a non-invasive measure of changes in blood volume, is widely used in both wearable devices and clinical settings. Recent PPG foundation models either use open-source ICU datasets with pretraining paradigms that require curated data and thus complicate generalization to field-like data, or use closed-source field-like PPG data. In contrast, we propose a PPG foundation model that does not require high-quality or field-like pretraining data, and instead leverages accompanying electrocardiogram and respiratory signals in ICU datasets to select contrastive samples during pretraining. Our approach allows the model to retain and learn from noisy PPG segments, improving robustness at inference. Our model, pretrained on 3x fewer subjects than existing state-of-the-art approaches, achieves performance improvements on 14 out of 15 diverse downstream tasks, including field-like daily activity and heart rate prediction. Our results demonstrate that multimodal supervision can integrate complementary physiological information to improve the robustness of PPG foundation models and enhance their generalization to consumer-grade data.
\end{abstract}

\begin{figure*}[!ht]
\begin{center}
\includegraphics[width=\textwidth]{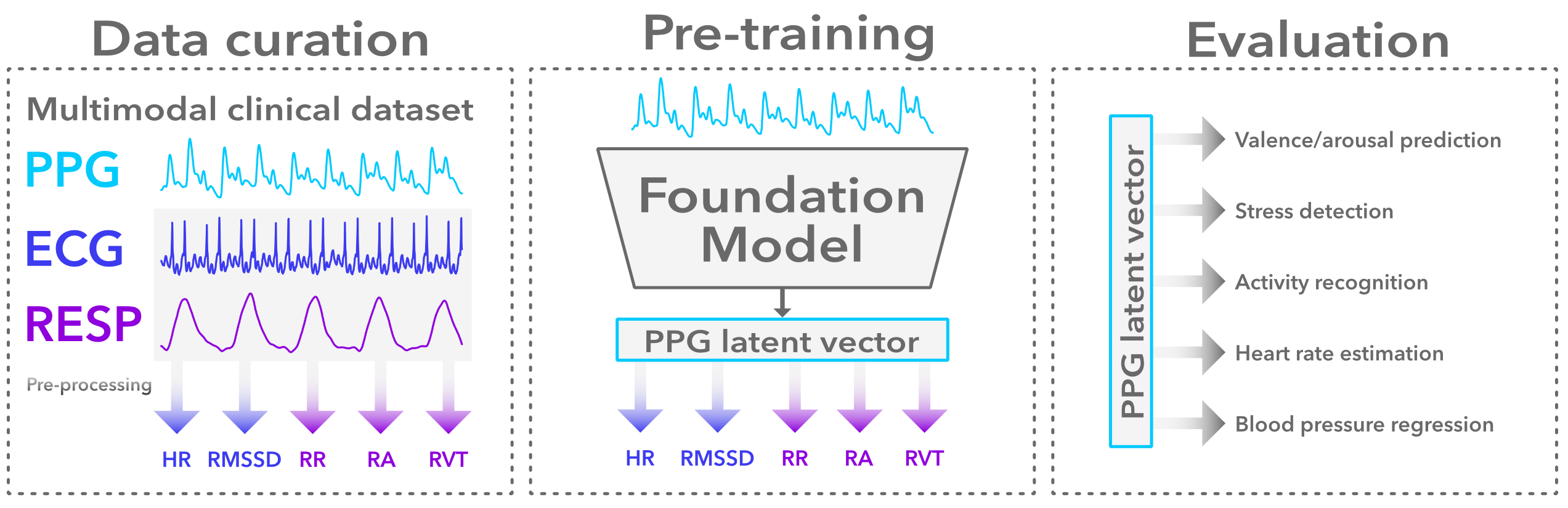}
\end{center}
\caption{\textbf{Multimodal contrastive supervision framework.} (Left) The electrocardiogram (ECG) and respiratory (RESP) data co-recorded with PPG is segmented into 10s windows. Five metrics are extracted from the ECG and RESP segments that summarize those windows in a 5-dimensional vector. (Middle) The metrics are used to generate contrastive samples during pretraining. (Right) The unimodal PPG embeddings are evaluated using various tasks for unseen datasets.}
\label{fig:framework}
\end{figure*}

\section{Introduction}
Wearable devices are rapidly emerging as powerful tools to monitor physiological and behavioral signals in everyday life. These devices typically rely on embedded sensors that must meet strict design constraints: they must be small, low-power, cost-effective, and unobtrusive. However, these constraints often compromise signal quality, introducing noise and variability that significantly challenge downstream tasks. Consequently, there is a critical need for robust models that can learn effective representations from noisy signals while maintaining high accuracy and generalizability across diverse conditions. Among the various biosignals used in wearables, photoplethysmography (PPG) has gained prominence due to its simplicity, low energy consumption, and compatibility with optically-based sensing hardware. PPG measures changes in blood volume in peripheral tissue, allowing the estimation of vital signs such as heart rate and blood pressure~\citep{elgendi2019use}. Unlike electrocardiography (ECG), PPG is more prone to motion artifacts and signal noise~\citep{fine2021sources}. Nevertheless, PPG’s ability to reflect vascular dynamics makes it a promising candidate for foundation models that aim to generalize across multiple tasks and conditions. 

The advent of PaPaGei~\citep{pillai2024papagei}, the first open-source PPG foundation model, highlights the growing interest in building general-purpose representations from wearable biosignals. PaPaGei demonstrates substantial performance gains over engineered features across multiple downstream tasks, including hypertension classification, blood pressure estimation, and heart rate prediction. These results suggest that foundation models can capture rich, transferable representations of PPG signals. However, PaPaGei depends on extensive preprocessing to extract clean morphological features for pretraining, and other existing approaches often rely on proprietary datasets~\citep{abbaspourazad2023large,saha2025pulse}, limiting reproducibility and/or scalability. In this work, we investigate whether the reliance on curated and denoised PPG signals can be relaxed by leveraging co-recorded multimodal signals from large-scale intensive care unit (ICU) datasets. Specifically, we propose a novel PPG foundation model that uses co-recorded high-quality signals (electrocardiogram (ECG) and respiratory (RESP) data) to select contrastive PPG samples during pretraining. This allows us to learn from relatively noisy clinical PPG data, without requiring explicit denoising, preprocessing, or morphological feature extraction. Importantly, our model only requires PPG signals at inference, with multimodal data used exclusively during pretraining to enhance representation learning, as shown in Figure~\ref{fig:framework}. Our key contributions are as follows:

\begin{itemize} 
\item We propose a multimodal-guided pretraining framework for PPG foundation models that leverages co-recorded physiological signals (ECG and RESP) as supervisory signals during pretraining, while requiring only PPG at inference time. This increases robustness to consumer-level noise without increasing deployment or sensing complexity.
\item We introduce within-subject linear probing as a complementary evaluation protocol for PPG foundation models, enabling fine-grained, subject-specific assessment of learned representations and revealing generalization properties beyond standard cross-subject evaluation. 
\item Extensive experiments across 15 downstream tasks show that our PPG foundation model outperforms five strong baselines on 14 out of 15 tasks, despite being pretrained on a single dataset with a substantially smaller subject pool. These results demonstrate the effectiveness, efficiency, and robustness of multimodal-guided pretraining, particularly for field-like deployment scenarios.
\end{itemize}

\paragraph{Conflict of Interest Disclosure}
The authors SD, GKV, LL, TM, and DD are employed by Dolby Laboratories, which leads the development of the model proposed in this paper, which was among the ones evaluated in this paper.

\section{Related work}
Foundation models have demonstrated strong generalization across a wide range of domains, driven by large-scale pretraining and self-supervised learning~\citep{bommasani2021opportunities}. These models are often trained using self-supervised learning techniques that involve generating masked or incomplete data~\citep{devlin2019bert}. Generative and/or predictive pretraining has been replicated with success in other fields as well, including computer vision~\citep{he2022masked}, pretraining for timeseries~\citep{nie2022time}, and biosignals~\citep{kostas2021bendr,chen2021forecasting,chien2022maeeg,dong2023simmtm,liu2023frequency,yun2024unsupervised,zhang2024general,geenjaar2025citrus}. An important downside to these methods is that they can be sensitive to noise. In practice, biosignals like PPG are far noisier than image or text data, and can be highly subject-dependent, influenced by factors such as skin tone and body composition~\citep{bent2020investigating}. Generative approaches may thus fail to capture good embeddings for high-noise segments. Predictive approaches like JEPA~\citep{lecun2022path} aim to mitigate this by learning abstract representations without full reconstruction. Still, they remain sensitive to slow-varying or predictive noise patterns~\citep{sobal2022jepa}. This slow and/or predictive noise may be induced by movement in PPG data. JEPA approaches may thus still struggle to learn good embeddings from noisy PPG data. Other, nongenerative or predictive self-supervised approaches have also been proposed for PPG data, including motif matching~\citep{xu2023rebar,xu2024relcon,saha2025pulse}, morphology-based contrastive learning~\citep{pillai2024papagei}, and temporal- or participant-based contrastive learning~\citep{tonekaboni2021unsupervised,abbaspourazad2023large}. Although these approaches do not rely on generation and/or prediction during pretraining, they may still be sensitive to noise. For instance, PaPaGei relies on morphological features that are difficult to extract accurately from noisy segments, which can result in poor supervision or the exclusion of data. Moreover, similar to JEPA approaches, time- or participant-based contrastive learning that uses two noisy positive pairs may learn to focus on slow-frequency features, as opposed to potentially important medium-frequency signals like systolic peaks in the PPG signal.

Unimodal self-supervised supervision is only one subset of foundation models for general representation learning~\citep{li2024multimodal}. Other subsets include label supervision, multimodal supervision, and multimodal fusion models~\citep{li2024multimodal}. Given that labels vary significantly between PPG datasets, and future inference may be restricted by training a PPG foundation model using a specific set of classes, it is not practical to use label supervision for PPG foundation models. Moreover, although multimodal biosignal foundation models can be made robust to modality dropout~\citep{liu2023frequency,fang2024promoting}, it is preferable to train a PPG foundation model that only requires PPG as an input to make it as general for wearable use as possible. Specifically, a model that can be trained using multimodal inputs but deployed using a single modality allows it to operate on a wider range of wearable devices that may not support all pretraining modalities. Multimodal pretraining is particularly relevant because large-scale open-source PPG pretraining datasets are often derived from clinical studies, where multiple physiological modalities are routinely recorded from patients as part of standard monitoring. To leverage the availability of such data, we therefore focus on multimodal supervision as a method to train a PPG foundation model. Multimodal supervision models in computer vision include CLIP~\citep{radford2021learning,jia2021scaling}, while in biosignals related approaches include multimodal contrastive pretraining~\citep{raghu2022contrastive}, BioFAME~\citep{liu2023frequency}, SleepFM~\citep{thapa2024sleepfm}, cross-modal masked auto-encoding~\citep{fang2024promoting}, and CiTrus~\citep{geenjaar2025citrus}. These works commonly incorporate signals that are not available in most wearable devices, such as ECG, EOG, or EEG. For PPG specifically, foundation models using multimodal supervision include SensorLM~\citep{zhang2025sensorlm}, as well as approaches that jointly leverage PPG and accelerometer data~\citep{abbaspourazad2024wearable}. Both models are trained on closed-source datasets. In contrast, our approach uses open-source ICU data and relies only on PPG at inference time. We leverage biosignals co-recorded with PPG to construct a physiologically grounded supervision signal during pretraining, enabling robust representation learning from noisy clinical PPG data while maintaining scalability and reproducibility. This allows the community to build, evaluate, and extend upon our method by training on the same set of data. In fact, we share a list of the exact data files we use for our pretraining dataset can be found in the \href{https://figshare.com/articles/online_resource/Supplementary_material_for_the_paper_A_robust_PPG_foundation_model_using_multimodal_physiological_supervision_/32478918?file=65050821}{Supplementary Materials}.

\section{Methods}
\label{sec:methods}
Let $\mathcal{D} = \{(\text{PPG}, \text{ECG}, \text{RESP})^{n, s}\}_{n=1...N, s=1...S}$ denote the multimodal biosignal dataset, where \(\text{N}\) is the number of subjects and \(S\) is the number of sessions per subject. Each tuple \((\text{PPG}, \text{ECG}, \text{RESP})^{n,s}\) contains time-aligned signals from each modality of equal length. The continuous signals are segmented into non-overlapping 10-second windows following PaPaGei~\citep{pillai2024papagei}, \(\mathbf{w}^{n,s}_t = (\text{PPG}, \text{ECG}, \text{RESP})^{n,s}_{t:(t+10s)}\). The PPG segment \(\mathbf{x}^{n,s}_t = \mathbf{w}^{n,s}_{t, \text{PPG}}\) is used as the model input, while the corresponding ECG and RESP segments are utilized to compute physiological metrics that guide contrastive supervision.

\subsection{A Physiological Metric Space for Contrastive Supervision}
\label{sec:multimodal-supervision}
To ensure our PPG foundation model is robust to naturally occurring noise, such noise must be well-represented in its pretraining dataset. Previous work relied on morphological features directly extracted from PPG segments to define contrastive targets~\citep{pillai2024papagei}. Noisy segments, which are important to include in the pretraining dataset~\citep{saha2025pulse}, consequently either need to be discarded or lead to inaccurate contrastive targets during pretraining. By contrast, ECG and RESP signals are typically less noisy and provide unique and physiologically relevant information. To ensure that noise sources like motion artifacts, which often co-occur in all modalities, do not affect contrastive target selection, we filter the ECG and RESP signals. However, filtering the raw ECG or RESP signal is challenging because it is unclear how noisy ECG or RESP segments should be imputed or interpolated. To ensure that co-occurring noise does not affect the multimodal supervision signal, we extract heart rate, heart-rate variability, and respiratory metrics. This allows us to use the known physiological ranges of the metrics, like heart rate, to filter out any noise, and improve contrastive target selection. The metrics are computed for each 10-second segment to reflect the underlying cardio-respiratory state of the subject. 

\textbf{ECG-derived metrics.} We extract two cardiovascular targets from each 10s ECG waveform: heart rate (\textbf{HR}) and the root mean square of successive differences (\textbf{RMSSD}). These are metrics that have relatively good repeatability for short segments~\citep{schroeder2004repeatability,nussinovitch2011reliability,shaffer2017overview}. Both provide a measure of heart rate variability (HRV), whereas resting HR is an indicator of all-around fitness and even cardiovascular disease~\citep{fox2007resting}, RMSSD is sensitive to autonomic function and stress~\citep{kim2018stress}. Especially because we are able to filter the derived metrics before using them as contrastive targets, they are less sensitive to peripheral noise than PPG and serve as a physiological target for cardiac dynamics. 

\textbf{RESP-derived metrics.} We compute three respiratory features from each 10s RESP waveform: respiratory rate (\textbf{RR}), respiratory amplitude (\textbf{RA}), and respiratory volume per time (\textbf{RVT}). These metrics reflect different aspects of breathing behavior, such as rhythm and tidal volume, and can indicate stress or enhanced attention~\citep{widjaja2013cardiorespiratory}, or disorders~\citep{brinkman2026physiology}.

By precomputing and filtering these metrics to ensure physiological plausibility and reduce noise-induced artifacts, we obtain a stable multidimensional metric space. Contrastive relationships are then defined on the basis of similarity in this space rather than using potentially noisy PPG morphology. This design enables the model to learn representations that are better aligned with meaningful physiological variation, and more robust to naturally occurring PPG noise.

\textbf{Pretraining setup and learning objective.} Given a batch of \(\text{B}\) PPG segments, two augmented views are generated for each segment, yielding \(2\text{B}\) inputs. These are encoded using a shared convolutional neural network \(f_{\theta}(\cdot)\), producing embeddings \(\{\mathbf{v}_i \in \mathbb{R}^{512} \}_{i=1\ldots2\text{B}}\). Each embedding \(\mathbf{v}_i\) is associated with a physiological metric vector \(\mathbf{y}_i\).
For each anchor embedding \(\mathbf{v}_i\), the other embeddings \(\mathbf{v}_j\) (\(j \neq i\)) are ranked according to the distance between their physiological targets \(d(\mathbf{y}_i, \mathbf{y}_j)\) in the metric space. Embeddings corresponding to more physiologically similar segments are ranked higher. Formally, define the set
$\mathcal{S}_{i,j} = \{ \mathbf{v}_k \mid k \neq i, \; d(\mathbf{y}_i, \mathbf{y}_k) \geq d(\mathbf{y}_i, \mathbf{y}_j) \}$, which contains all embeddings that are further away than \(\mathbf{v}_j\) is from \(\mathbf{v}_i\). Then, we employ the rank-n-contrast (RNC) loss~\citep{zha2023rank}, which encourages embeddings that are closer in the physiological metric space to be closer in the learned representation space as well:
\begin{equation}
\label{eq:loss}
\begin{aligned}
\mathcal{L}_{\text{RNC}}
&= \frac{1}{2B(2B-1)}
   \sum_{i=1}^{2B}
   \sum_{\substack{j=1 \\ j \neq i}}^{2B}
   \ell(\mathbf{v}_i, \mathbf{v}_j),
\\
\ell(\mathbf{v}_i, \mathbf{v}_j)
&= -\log
   \frac{\exp\!\left(\mathrm{sim}(\mathbf{v}_i, \mathbf{v}_j)/\tau\right)}
        {\sum_{\mathbf{v}_k \in \mathcal{S}_{i,j}}
         \exp\!\left(\mathrm{sim}(\mathbf{v}_i, \mathbf{v}_k)/\tau\right)} .
\end{aligned}
\end{equation}
where \(\text{sim}(\cdot, \cdot)\) denotes cosine similarity and \(\tau\) is a temperature hyperparameter. This loss anchors the learned PPG embeddings to the robust physiological metric space derived from multimodal signals, improving noise robustness and encouraging physiologically meaningful representations.

\textbf{Unimodal inference.} Although we use co-recorded biosignals in PPG model pretraining, we would like to emphasize that our model only uses PPG during inference, making it applicable for a wide range of consumer-level applications.

\begin{table}[t]
\centering
\scriptsize
\setlength{\tabcolsep}{3.5pt} 
\renewcommand{\arraystretch}{1.05}

\caption{\textbf{Downstream dataset information}. More information is provided in Appendix~\ref{app:downstream-datasets}. \# P is the number of participants, \# S is the  number of total samples, HR is heart rate, and IR is infrared.}
\label{tab:datasets}

\begin{tabularx}{\columnwidth}{lXccc}
\hline
Datasets & Task & Task type & \# P & \# S \\
\hline
WESAD & Stress & Clf (2) & 15 & 4125 \\
\citep{schmidt2018introducing} & Affect & Clf (4) & 15 & 4125 \\
\hline
PPG-DaLiA & Daily \linebreak activities & Clf (9) & 15 & 12865 \\
\citep{reiss2019deep} & HR & Reg & 15 & 64697 \\
\hline
EEVR & Valence & Clf (2) & 37 & 10508 \\
\citep{singh2024eevr} & Arousal & Clf (2) & 37 & 10508 \\
\hline
PPG-BP & Hypertension & Clf (2) & 219 & 657 \\
 & Average HR & Reg & 219 & 657 \\
\multirow{3}{*}{\citep{liang2018new}} & Systolic BP & Reg & 219 & 657 \\
 & Diastolic BP & Reg & 219 & 657 \\
\hline
VitalVideos & Systolic BP & Reg & 100 & 300 \\
\citep{toye2023vital} & Diastolic BP & Reg & 100 & 300 \\
\hline
WildPPG & HR (green) & Reg & 64 & 304708 \\
\multirow{2}{*}{\citep{meier2024wildppg}} & HR (IR) & Reg & 64 & 304708 \\
 & HR (red) & Reg & 64 & 304708 \\
\hline
\end{tabularx}
\end{table}

\section{Experimental Setup}
\textbf{Datasets.}
For pretraining, we use the MIMIC-III Waveform Database Matched Subset~\citep{goldberger2000physiobank,johnson2016mimic,moody2020mimic}\footnote{\url{https://physionet.org/content/mimic3wdb-matched/1.0/}}, 
which contains waveform data from 10,282 ICU patients. We selected this dataset among the three used by PaPaGei because its subjects are neither asleep nor under anesthesia, unlike the other two datasets. This allows patients to move their arms naturally, introducing realistic movement artifacts and noise that improve the robustness of our model. The dataset includes multiple time-aligned biosignals sampled at 125Hz. The pretraining data preprocessing pipeline, detailed in Appendix~\ref{app:multimodal-preprocessing}, filters the data to retain 4,998 subjects, yielding approximately 20 million 10-second PPG segments (about 56,000 hours of data). We evaluate our model on unseen datasets and tasks, an overview of these datasets is given in Table~\ref{tab:datasets}. In our selection of downstream datasets we focus on wearable-level data to verify the robustness of our model to noise in the PPG signal. Specifically, PPG-BP and VitalVideos are similar to clinical-level PPG data, WESAD and EEVR are lab environment datasets, and DaLiA and WildPPG are field-like datasets, which exhibit the highest noise levels.

\textbf{Backbone \& pretraining.}
We use a 1D ResNet-26 convolutional encoder~\footnote{\url{https://github.com/hsd1503/resnet1d}} \(f_{\theta}(\cdot)\) with instance normalization applied to the input. The network comprises 12 residual blocks, each using a kernel size of 11 and stride 2. The initial convolution outputs 128 filters, doubling every four layers. Spatial resolution is downsampled by a factor of 2 every two layers via max-pooling. Each unfiltered 10-second PPG window \(\mathbf{x}^{n, t}\) is passed through the network to produce embeddings used in our contrastive learning objective (Eq.~\ref{eq:loss}). For data augmentation, two random transformations are applied to each input window, selected from: \texttt{GaussianNoise} (p = 0.25), \texttt{Negation} (p = 0.20), \texttt{Scaling} (p = 0.40), and \texttt{RandomCrop} (p = 0.50). These augmentations follow the same strategy as PaPaGei~\citep{pillai2024papagei}. Additional hyperparameter details are provided in Appendix~\ref{app:hyperparameters}. 

\begin{figure*}[!ht]
\begin{center}
\includegraphics[width=\textwidth]{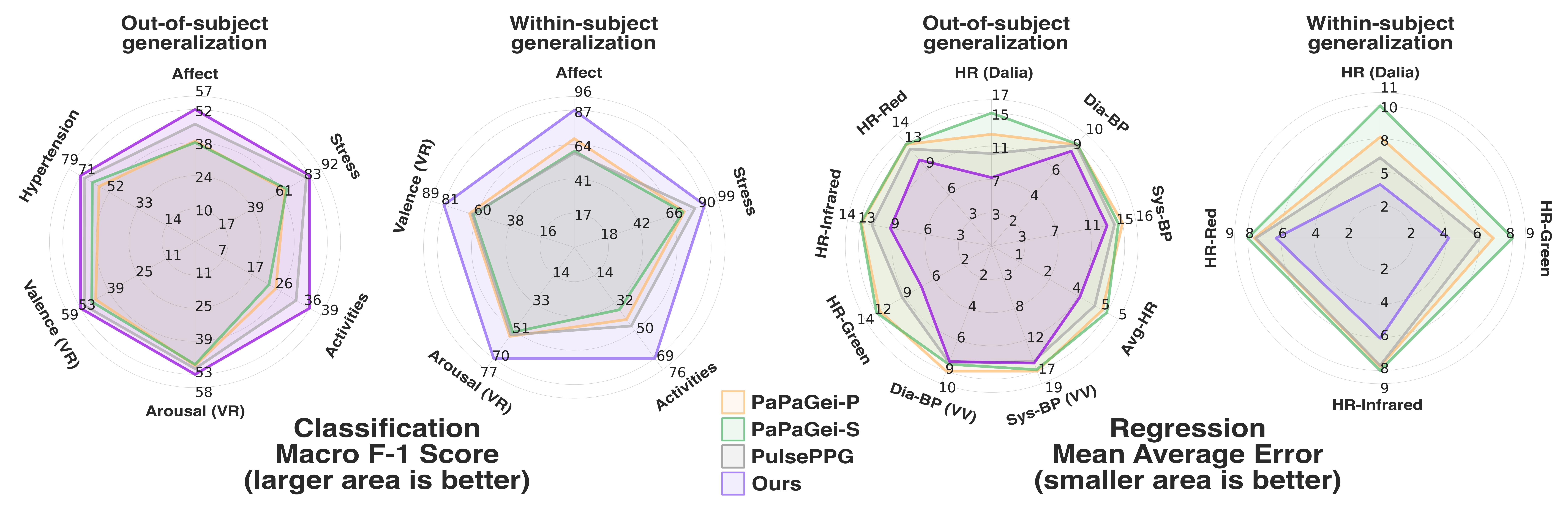}
\end{center}
\caption{\textbf{Comparison with state-of-the-art}. (Left) Classification results in terms of their macro F-1 score (\textbf{larger area is better}). (Right) Regression results in terms of their mean average error (\textbf{smaller area is better}). We evaluate across subject linear probing, and within subject linear probing.}
\label{fig:overview-results}
\end{figure*}

\textbf{Evaluation across subjects.}
We evaluate our foundation model on 15 downstream tasks from six unseen datasets, encompassing both classification and regression problems. Classification tasks include stress, affect, arousal, valence, activity, and hypertension detection, while regression tasks cover heart rate prediction in field, daily activity, and clinical settings, as well as diastolic and systolic blood pressure estimation. Table~\ref{tab:datasets} provides a detailed list of datasets and tasks, with more information in Appendix~\ref{app:downstream-datasets}. Following PaPaGei~\citep{pillai2024papagei}, we assess representation quality and generalizability using linear probing. Linear probing measures linear predictability from inferred embeddings while keeping the backbone weights frozen. For classification tasks, we use logistic regression, and for regression we use ridge regression. Hyperparameters for both models are tuned via 5-fold cross-validation on the training and validation splits, and are discussed in Appendix~\ref{app:hyperparameters}. Both probes are implemented using \texttt{scikit-learn}~\citep{scikit-learn}. Final results are averaged over five test folds. Model selection uses macro F1 score for classification and mean absolute error (MAE) for regression. Additionally, we report accuracy (ACC) and area under the receiver operating characteristic curve (AUC) for classification, as well as mean squared error (MSE) and mean absolute percentage error (MAPE) for regression.

\textbf{Evaluation within subjects.}
To better assess performance in realistic deployment settings, we introduce a within-subject linear probing protocol. Wearable devices are typically used by individual users, and labeled data can often be obtained over time via user interaction or automatic annotation. Because physiological patterns vary substantially across individuals, evaluating linear probe performance separately per subject provides insight into generalization under subject-specific distributions. We use the same linear probe architecture, hyperparameter tuning strategy, and metrics as in across-subject evaluation. However, each fold corresponds to a single user rather than a subset of subjects. Training, validation, and test splits are constructed from the temporally varying targets of that user, and results are averaged over five randomly selected users (details in Appendix~\ref{app:hyperparameters}). Only datasets with temporally varying per-subject targets are included (e.g., affect detection and heart rate), while datasets with static labels, such as PPG-BP, are excluded, yielding nine tasks from the WESAD, DaLiA, EEVR, and WildPPG datasets.

\textbf{Baselines.} We compare our model against PaPaGei-S, PaPaGei-P~\citep{pillai2024papagei}, SimCLR, BYOL, and PulsePPG~\citep{saha2025pulse}. When available, we use the official open-source weights for each model. Comparisons with PaPaGei-S, PaPaGei-P, and PulsePPG are presented in the main text, while across-subject results for BYOL and SimCLR are reported in Appendix~\ref{app:baselines}. For SimCLR and BYOL, we use the same backbone architecture, model size, and pre-training data as our proposed model to ensure a controlled and fair comparison. In addition to domain-specific foundation models like PaPaGei and PulsePPG, we also compare our model against Chronos and MOMENT, which are general-purpose timeseries foundation models. For Chronos, we use the Chronos-T5-base (200M) backbone~\cite{ansari2024chronos}, and for MOMENT we use MOMENT-1-Large (385M)~\cite{goswami2024moment}. In both cases, we use the publicly available checkpoints, and evaluate the model embeddings on our downstream evaluation data splits. The results for these general-purpose foundation models are reported in Appendix~\ref{app:foundation-model}.

\begin{table*}[!ht]
\centering
\small
\setlength{\tabcolsep}{3.5pt} 
\renewcommand{\arraystretch}{1.05}

\caption{\textbf{Downstream across subjects linear probing results}. Results for each method are averaged across $5$ test folds, and standard deviations can be found in Appendix~\ref{app:standard-deviations}. For the classification tasks, higher is better, and evaluation metrics are macro F-1 score (MF1), accuracy (ACC), and the area under the receiver operating characteristic (AUC). For regression tasks lower is better, and we use mean average error (MAE), mean squared error (MSE), and mean average percentage error (MAPE).} 
\begin{tabular}{l|xxx|yyy|ggg|zzz|}
 & \multicolumn{3}{c}{\cellcolor{PPGPTable}\textbf{PaPaGei-P}} & \multicolumn{3}{c}{\cellcolor{PPGSTable}\textbf{PaPaGei-S}} & \multicolumn{3}{c}{\cellcolor{Gray}\textbf{PulsePPG}} & \multicolumn{3}{c}{\cellcolor{OURTable}\textbf{Ours}} \\ 
\hline 
Clf ($\uparrow$) & \textbf{MF1} & \textbf{ACC} & \textbf{AUC} & \textbf{MF1} & \textbf{ACC} & \textbf{AUC} & \textbf{MF1} & \textbf{ACC} & \textbf{AUC} & \textbf{MF1} & \textbf{ACC} & \textbf{AUC} \\ 
\hline 
Stress &  0.65 &  0.79 &  0.8 &  0.67 &  0.78 &  0.76 &  0.81 &  0.86 &  0.92 &  0.83 &  0.88 &  0.94 \\ 
Affect &  0.4 &  0.49 &  0.69 &  0.39 &  0.47 &  0.66 &  0.46 &  0.55 &  0.74 &  0.52 &  0.6 &  0.78 \\ 
Activities &  0.25 &  0.33 &  0.72 &  0.23 &  0.29 &  0.68 &  0.31 &  0.38 &  0.79 &  0.36 &  0.41 &  0.82 \\ 
Arousal &  0.49 &  0.56 &  0.53 &  0.49 &  0.54 &  0.51 &  0.51 &  0.56 &  0.53 &  0.53 &  0.57 &  0.57 \\ 
Valence &  0.46 &  0.63 &  0.55 &  0.48 &  0.62 &  0.54 &  0.51 &  0.62 &  0.57 &  0.53 &  0.62 &  0.58 \\ 
Hypertension &  0.6 &  0.65 &  0.64 &  0.64 &  0.69 &  0.71 &  0.69 &  0.73 &  0.74 &  0.71 &  0.75 &  0.77 \\ 
\hline 
\rowcolor{white} 
Avg & 0.48  & 0.58  & 0.65  & 0.48  & 0.56  & 0.64  & 0.55  & 0.62  & 0.71  & \textbf{0.58}  & \textbf{0.64}  & \textbf{0.74}  \\ 
\hline 
Reg ($\downarrow$) & \textbf{MAE} & \textbf{MSE} & \textbf{MAPE} & \textbf{MAE} & \textbf{MSE} & \textbf{MAPE} & \textbf{MAE} & \textbf{MSE} & \textbf{MAPE} & \textbf{MAE} & \textbf{MSE} & \textbf{MAPE} \\ 
\hline 
HR (Dalia) &  12.7 &  303 &  0.14 &  15.1 &  407 &  0.17 &  10.5 &  221 &  0.12 &  7.78 &  143 &  0.09 \\ 
Avg-HR &  4.76 &  40.0 &  0.07 &  4.95 &  43.4 &  0.07 &  4.43 &  34.4 &  0.06 &  3.8 &  26.3 &  0.05 \\ 
Sys-BP &  14.9 &  366 &  0.12 &  14.4 &  351 &  0.12 &  14.0 &  336 &  0.11 &  13.2 &  281 &  0.11 \\ 
Dia-BP &  8.73 &  121 &  0.12 &  8.77 &  122 &  0.12 &  8.68 &  121 &  0.12 &  8.16 &  109 &  0.12 \\ 
Sys-BP (VV) &  17.0 &  516 &  0.13 &  16.8 &  500 &  0.13 &  15.7 &  439 &  0.12 &  15.9 &  451 &  0.12 \\ 
Dia-BP (VV) &  8.7 &  143 &  0.11 &  8.22 &  125 &  0.1 &  8.12 &  119 &  0.1 &  8.04 &  123 &  0.1 \\ 
HR-Green &  12.2 &  266 &  0.17 &  12.5 &  273 &  0.17 &  9.68 &  188 &  0.13 &  7.61 &  149 &  0.1 \\ 
HR-Infrared &  12.6 &  273 &  0.17 &  12.7 &  277 &  0.17 &  11.6 &  246 &  0.16 &  9.82 &  206 &  0.14 \\ 
HR-Red &  12.7 &  279 &  0.17 &  12.7 &  284 &  0.17 &  12.1 &  261 &  0.17 &  10.8 &  231 &  0.15 \\ 
\hline 
\rowcolor{white} 
Avg & 11.6  & 256  & 0.13  & 11.8  & 265  & 0.14  & 10.5  & 218  & 0.12  & \textbf{9.45}  & \textbf{191}  & \textbf{0.11}  \\ 
\hline 
\end{tabular} 
\label{tab:linear-probing-group} 

\end{table*}
\begin{table*}
\centering
\small
\setlength{\tabcolsep}{3.5pt} 
\renewcommand{\arraystretch}{1.05}

\caption{\textbf{Downstream within subjects linear probing results}. Results for each method are averaged across $5$ test folds, and standard deviations can be found in Appendix~\ref{app:standard-deviations}. For the classification tasks, higher is better, and evaluation metrics are macro F-1 score (MF1), accuracy (ACC), and the area under the receiver operating characteristic (AUC). For regression tasks lower is better, and we use mean average error (MAE), mean squared error (MSE), and mean average percentage error (MAPE).} 
\begin{tabular}{l|xxx|yyy|ggg|zzz|}
 & \multicolumn{3}{c}{\cellcolor{PPGPTable}\textbf{PaPaGei-P}} & \multicolumn{3}{c}{\cellcolor{PPGSTable}\textbf{PaPaGei-S}} & \multicolumn{3}{c}{\cellcolor{Gray}\textbf{PulsePPG}} & \multicolumn{3}{c}{\cellcolor{OURTable}\textbf{Ours}} \\ 
\hline 
Clf ($\uparrow$) & \textbf{MF1} & \textbf{ACC} & \textbf{AUC} & \textbf{MF1} & \textbf{ACC} & \textbf{AUC} & \textbf{MF1} & \textbf{ACC} & \textbf{AUC} & \textbf{MF1} & \textbf{ACC} & \textbf{AUC} \\ 
\hline 
Stress &  0.76 &  0.83 &  0.76 &  0.75 &  0.81 &  0.77 &  0.83 &  0.86 &  0.92 &  0.9 &  0.92 &  0.96 \\ 
Affect &  0.69 &  0.73 &  0.83 &  0.61 &  0.66 &  0.83 &  0.6 &  0.67 &  0.87 &  0.87 &  0.89 &  0.96 \\ 
Activities &  0.45 &  0.51 &  0.82 &  0.39 &  0.44 &  0.75 &  0.49 &  0.52 &  0.87 &  0.69 &  0.69 &  0.94 \\ 
Arousal &  0.56 &  0.59 &  0.61 &  0.54 &  0.56 &  0.58 &  0.55 &  0.58 &  0.61 &  0.7 &  0.71 &  0.76 \\ 
Valence &  0.65 &  0.69 &  0.69 &  0.63 &  0.67 &  0.66 &  0.64 &  0.67 &  0.7 &  0.81 &  0.83 &  0.86 \\ 
\hline 
\rowcolor{white} 
Avg & 0.62  & 0.67  & 0.74  & 0.58  & 0.63  & 0.72  & 0.62  & 0.66  & 0.79  & \textbf{0.79}  & \textbf{0.81}  & \textbf{0.9}  \\ 
\hline 
Reg ($\downarrow$) & \textbf{MAE} & \textbf{MSE} & \textbf{MAPE} & \textbf{MAE} & \textbf{MSE} & \textbf{MAPE} & \textbf{MAE} & \textbf{MSE} & \textbf{MAPE} & \textbf{MAE} & \textbf{MSE} & \textbf{MAPE} \\ 
\hline 
HR (Dalia) &  7.92 &  121 &  0.09 &  10.4 &  199 &  0.12 &  6.29 &  82.7 &  0.07 &  4.2 &  40.0 &  0.05 \\ 
HR-Green &  7.22 &  122 &  0.09 &  8.45 &  186 &  0.11 &  6.35 &  97.8 &  0.08 &  4.38 &  65.9 &  0.06 \\ 
HR-Infrared &  8.03 &  142 &  0.11 &  8.27 &  151 &  0.11 &  7.95 &  143 &  0.11 &  6.25 &  107 &  0.08 \\ 
HR-Red &  7.85 &  130 &  0.1 &  8.26 &  185 &  0.11 &  7.76 &  128 &  0.1 &  6.5 &  103 &  0.08 \\ 
\hline 
\rowcolor{white} 
Avg & 7.76  & 129  & 0.1  & 8.83  & 180  & 0.11  & 7.09  & 113  & 0.09  & \textbf{5.33}  & \textbf{78.9}  & \textbf{0.07}  \\ 
\hline 
\end{tabular} 
\label{tab:linear-probing-subject} 

\end{table*}
\section{Results}
Our model outperforms all baselines on nearly all (\textbf{14 out of 15}) downstream tasks in across-subject and (9 out of 9) within-subject linear probing evaluations, as shown in Figure~\ref{fig:overview-results}. Notably, these gains are achieved despite using \textbf{3x} fewer pretraining subjects than PaPaGei, and without requiring field-like pretraining data as in prior work, PulsePPG~\cite{saha2025pulse}. This demonstrates the efficiency and robustness of our approach. Detailed numerical results for the across-subject evaluation are provided in Table~\ref{tab:linear-probing-group}, while within-subject results are summarized in Table~\ref{tab:linear-probing-subject}. These results highlight the effectiveness and versatility of our model across diverse PPG analysis scenarios.

\textbf{Evaluation across subjects.} Our model achieves consistent and substantial improvements over the PaPaGei and PulsePPG models. Notable classification gains include stress (\textbf{0.83 vs.\ 0.81}), affect detection (\textbf{0.52 vs.\ 0.46}), and hypertension classification (\textbf{0.71 vs.\ 0.69}) reflecting stronger generalization across user states, and generalization to lab and clinical data. Regression tasks also show large improvements both under field-like (DaLiA: \textbf{7.78 vs.\ 10.5}, WildPPG (Green): \textbf{7.61 vs.\ 9.68}) and clinical conditions (Sys-BP: \textbf{13.2 vs.\ 14.0}). The only tasks our model underperforms PulsePPG at is systolic blood pressure regression for the VitalVideos dataset (\textbf{15.9 vs.\ 15.7}). This may be due to subtle differences in the pretraining data. However, our model performs better in terms of diastolic blood pressure regression for the same dataset (\textbf{8.04 vs.\ 8.12}). Our results replicate the general improvement of PaPaGei-S over PaPaGei-P, and the improvement of the PulsePPG model especially on field-like datasets as reported in the PaPaGei~\cite{pillai2024papagei} and PulsePPG papers~\cite{saha2025pulse}. Minor differences in our paper's absolute numbers stem from our use of K-fold cross-validation, which better captures subject variability compared to single-split setups~\citep{geenjaar2025citrus}.

\textbf{Evaluation within subjects.} Similar to the across subjects results, we find that classification improvements within subjects are also higher for stress (\textbf{0.9 vs.\ 0.83}), affect (\textbf{0.87 vs.\ 0.69}), and daily activities classification (\textbf{0.69 vs.\ 0.49}). We find that our model outperforms every model on within-subject classification and regression tasks. Moreover, we see a large increase in performance for valence and arousal detection, which could highlight that they are highly unique to each subject. Further, Appendix~\ref{app:within-subject} shows that even when trained with only 10\% of the available data, our model consistently outperforms all baseline models trained with the full dataset. Moreover, Appendix~\ref{app:within-subject} also provides visualizations of within-subject embedding structures, offering additional insight into the quality of the learned representations.

\textbf{Unimodal vs multimodal pretraining ablation.} To verify that differences in performance are due to our use of multimodal contrastive guidance during pretraining, and not because of architectural and pretraining data differences, we pretrain PaPaGei-S based on the available code~\footnote{\url{https://github.com/Nokia-Bell-Labs/papagei-foundation-model}} on our data. We adopt the PaPaGei-S pretraining objective, as it was shown to consistently outperform PaPaGei-P across downstream tasks in the original work. As a comparison, we match PaPaGei's backbone in our model, and use our proposed multimodal pretraining. For a fair comparison, we match the pretraining data, and all training hyperparameters between the models, differing only in the pretraining objective. As detailed in Table~\ref{tab:ablation-unimodal} in Appendix~\ref{app:ablation:pretraining}, our multimodal pretraining consistently and substantially outperforms unimodal pretraining across all evaluated tasks, except diastolic blood pressure regression. These results strongly validate our core hypothesis: integrating complementary biosignal modalities during contrastive learning effectively mitigates the limitations inherent in unimodal morphology-based contrastive targets, leading to significantly enhanced robustness, generalization, and downstream task performance. To further assess our model's sensitivity to noise, we perform an additional analysis on the PPG-DaLiA dataset in Appendix~\ref{app:noise}.

\paragraph{Scaling ablation.}
The PaPaGei paper shows that increasing its backbone model parameters does not necessarily lead to improvements across downstream tasks. To verify that our model's performance scales well with backbone size, we compare the PaPaGei backbone with our proposed backbone in Appendix~\ref{app:ablation:architecture}. The proposed larger backbone improves over the PaPaGei backbone on all but the average heart rate estimation, which indicates that our pretraining method likely benefits from further model size scaling. Contrastive learning and projector ablations are presented in Appendices~\ref{app:regression-ablation} and~\ref{app:projector-ablation}.
\begin{table*}[!ht] 
\centering
\small
\setlength{\tabcolsep}{3.5pt}       
\renewcommand{\arraystretch}{1.05}  

\caption{\textbf{ECG vs RESP ablation}. Results for each method are averaged across $5$ test folds, and standard deviations can be found in Appendix~\ref{app:standard-deviations}. For the classification tasks, higher is better, and evaluation metrics are macro F-1 score (MF1), accuracy (ACC), and the area under the receiver operating characteristic (AUC). For regression tasks lower is better, and we use mean average error (MAE), mean squared error (MSE), and mean average percentage error (MAPE).} 
\begin{tabular}{l|ggg|ggg|ggg|zzz|}
 & \multicolumn{3}{c}{\cellcolor{Gray}\textbf{HR + RESP}} & \multicolumn{3}{c}{\cellcolor{Gray}\textbf{ECG}} & \multicolumn{3}{c}{\cellcolor{Gray}\textbf{RESP}} & \multicolumn{3}{c}{\cellcolor{OURTable}\textbf{ECG + RESP}} \\ 
\hline 
Clf ($\uparrow$) & \textbf{MF1} & \textbf{ACC} & \textbf{AUC} & \textbf{MF1} & \textbf{ACC} & \textbf{AUC} & \textbf{MF1} & \textbf{ACC} & \textbf{AUC} & \textbf{MF1} & \textbf{ACC} & \textbf{AUC} \\ 
\hline 
Stress &  0.83 &  0.88 &  0.94 &  0.83 &  0.88 &  0.94 &  0.83 &  0.88 &  0.94 &  0.83 &  0.88 &  0.94 \\ 
Affect &  0.52 &  0.6 &  0.77 &  0.53 &  0.61 &  0.78 &  0.52 &  0.6 &  0.77 &  0.52 &  0.6 &  0.78 \\ 
Activities &  0.36 &  0.41 &  0.82 &  0.36 &  0.41 &  0.82 &  0.36 &  0.41 &  0.81 &  0.36 &  0.41 &  0.82 \\ 
Arousal &  0.53 &  0.56 &  0.55 &  0.52 &  0.54 &  0.53 &  0.53 &  0.57 &  0.56 &  0.53 &  0.57 &  0.57 \\ 
Valence &  0.54 &  0.61 &  0.58 &  0.54 &  0.61 &  0.57 &  0.54 &  0.61 &  0.58 &  0.53 &  0.62 &  0.58 \\ 
Hypertension &  0.71 &  0.74 &  0.76 &  0.7 &  0.73 &  0.76 &  0.7 &  0.73 &  0.75 &  0.71 &  0.75 &  0.77 \\ 
\hline 
\rowcolor{white} 
Avg & \textbf{0.58}  & 0.63  & 0.74  & 0.58  & 0.63  & 0.73  & 0.58  & 0.63  & 0.74  & 0.58  & \textbf{0.64}  & \textbf{0.74}  \\ 
\hline 
Reg ($\downarrow$) & \textbf{MAE} & \textbf{MSE} & \textbf{MAPE} & \textbf{MAE} & \textbf{MSE} & \textbf{MAPE} & \textbf{MAE} & \textbf{MSE} & \textbf{MAPE} & \textbf{MAE} & \textbf{MSE} & \textbf{MAPE} \\ 
\hline 
HR (Dalia) &  8.29 &  153 &  0.09 &  6.9 &  123 &  0.08 &  8.97 &  170 &  0.1 &  7.78 &  143 &  0.09 \\ 
Avg-HR &  4.14 &  28.6 &  0.06 &  3.66 &  23.2 &  0.05 &  4.06 &  28.0 &  0.06 &  3.8 &  26.3 &  0.05 \\ 
Sys-BP &  13.3 &  289 &  0.11 &  13.2 &  289 &  0.11 &  13.4 &  288 &  0.11 &  13.2 &  281 &  0.11 \\ 
Dia-BP &  8.18 &  109 &  0.12 &  8.38 &  113 &  0.12 &  8.3 &  110 &  0.12 &  8.16 &  109 &  0.12 \\ 
Sys-BP (VV) &  15.9 &  454 &  0.12 &  15.9 &  457 &  0.12 &  15.5 &  438 &  0.12 &  15.9 &  451 &  0.12 \\ 
Dia-BP (VV) &  8.13 &  120 &  0.1 &  8.33 &  130 &  0.1 &  8.11 &  116 &  0.1 &  8.04 &  123 &  0.1 \\ 
HR-Green &  7.99 &  154 &  0.11 &  7.17 &  142 &  0.1 &  8.53 &  167 &  0.12 &  7.61 &  149 &  0.1 \\ 
HR-Infrared &  10.2 &  214 &  0.14 &  9.38 &  195 &  0.13 &  10.6 &  222 &  0.15 &  9.82 &  206 &  0.14 \\ 
HR-Red &  11.2 &  241 &  0.15 &  10.4 &  222 &  0.14 &  11.4 &  246 &  0.16 &  10.8 &  231 &  0.15 \\ 
\hline 
\rowcolor{white} 
Avg & 9.7  & 196  & 0.11  & \textbf{9.25}  & \textbf{188}  & \textbf{0.1}  & 9.87  & 198  & 0.11  & 9.45  & 191  & 0.11  \\ 
\hline 
\end{tabular} 
\label{tab:linear-probing-metrics}
\end{table*}

\textbf{Metric ablation.} To assess the contribution of different supervisory metrics, we compare models pretrained using various combinations of auxiliary signals, including ECG-only, RESP-only, and ECG+RESP supervision. Detailed results are reported in Table~\ref{tab:linear-probing-metrics}. Overall, both ECG-guided and RESP-guided pretraining yield strong performance across downstream tasks. ECG-only supervision achieves the best average performance on regression tasks, whereas combined ECG+RESP supervision performs best on average for classification tasks. Notably, the ECG+RESP model outperforms baseline methods on the largest number of downstream tasks (14 out of 15), and is therefore selected as the most versatile and robust configuration. Moreover, supervision from both ECG and RESP improves generalization even for downstream tasks that are not directly related to these metrics. For instance, RESP-only pretraining using our proposed framework surpasses all baselines on field-like heart-rate regression benchmarks (Dalia and WildPPG). Moreover, ECG-only pretraining surpasses all baselines on systolic blood pressure regression, demonstrating that the observed performance gains primarily arise from the proposed multimodal-guided pretraining methodology rather than from direct correlations between the supervisory signals and downstream tasks.

\begin{figure*}
\centering
\begin{minipage}{0.49\textwidth}
    \centering    \includegraphics[width=\linewidth]{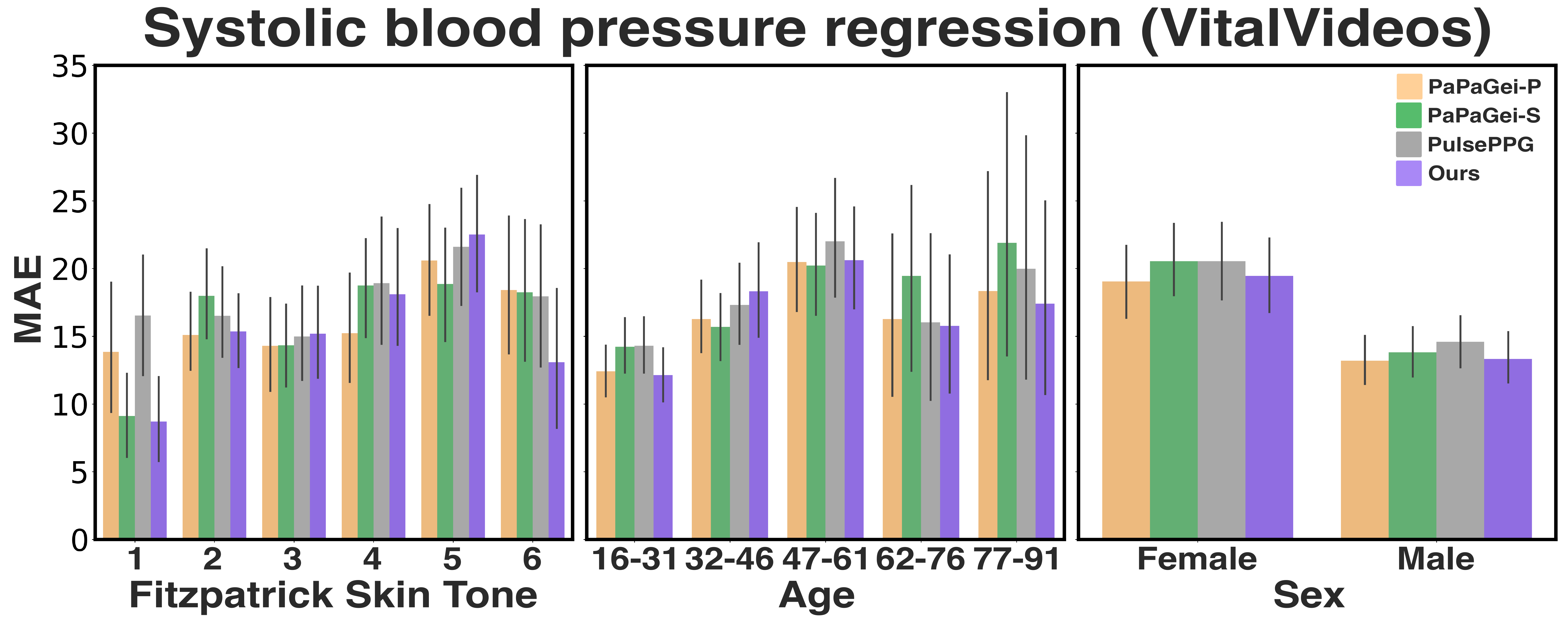}
    \caption{\textbf{Systolic blood pressure regression comparison across demographic variables.}}
    \label{fig:demographics}
\end{minipage}\hfill
\begin{minipage}{0.49\textwidth}
    \centering
    \includegraphics[width=\linewidth]{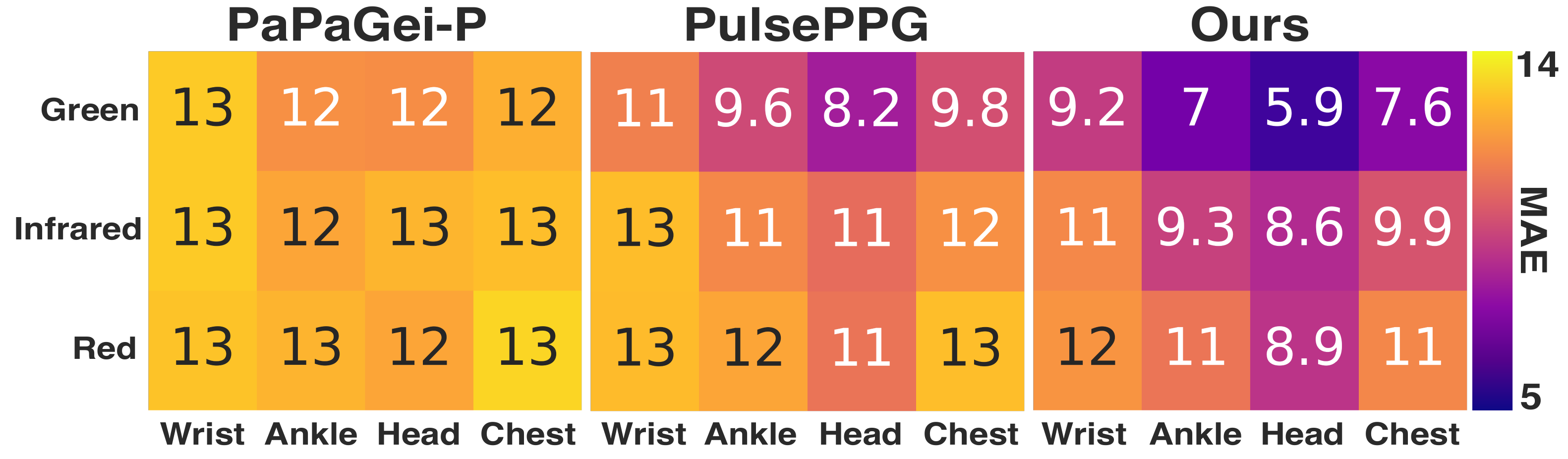}
    \caption{\textbf{WildPPG heart rate estimation comparison across PPG sensor location (x-axis) and type (y-axis).}}
    \label{fig:wild-results}
\end{minipage}
\end{figure*}

\textbf{Demographic analysis.} It is important to ensure that neither the specific set of subjects in the pretraining dataset nor the pretraining method lead to demographic biases in the model. To check for any demographic biases, we use the systolic blood pressure regression task on the VitalVideos dataset, which records the Fitzpatrick skin tone~\citep{gupta2019skin}, age, and sex of each subject. We use a leave-one-subject-out approach to perform linear probing, and also to select hyperparameters on the training set. The importance of skin tone in PPG analyses cannot be understated because PPG is an optical method, and skin tone can affect light wave reflectance~\citep{fallow2013influence}. Moreover, general device error for wearables recording heart rate has been found to be higher for darker skin tones~\citep{gupta2019skin}. The results in Figure~\ref{fig:demographics} show that some of the known biases appear in the models we tested. Specifically, performance is best for skin tone $1$, and in case of skin tone $5$ we see that our model performs worse than the the baseline models, but better for skin tone $6$. In addition, performance for adults aged 47-61 is worst and performance is lower for female subjects. These findings underscore existing challenges in equitable biosignal modeling and highlight areas for future bias mitigation.

\textbf{Heart rate estimation ablation.} Figure~\ref{fig:wild-results} characterizes heart rate estimation performance in field-like conditions across different PPG recording locations and sensor types (green, red, and infrared). We compare against PaPaGei-P instead of PaPaGei-S because it performs better on average for the WildPPG heart rate estimation tasks. Our model consistently achieves the best performance across all conditions, with particularly strong results for green wrist-worn PPG sensors. It is also clear from these results that all models perform best for head-based PPG sensors. An additional ablation experiment that compares heart rate regression performance with NeuroKit’s~\citep{Makowski2021neurokit} automatic heart rate estimation tool is presented in Appendix~\ref{app:neurokit}.

\section{Discussion}
This paper presents a PPG foundation model with a robust multimodal pretraining pipeline that utilizes multiple biosignals alongside PPG to guide contrastive foundation model pretraining. By extracting metrics from co-recorded biosignals, we leverage their known physiological bounds to filter noise and construct reliable contrastive targets. This improves robustness beyond clinical data and leads to improved downstream performance. Through ablation studies, we show consistent gains, with our model outperforming all baselines on all but two of 15 downstream tasks in the across-subject setting, and on all tasks in the within-subject setting. Notably, it achieves large improvements on field-like datasets (DaLiA and WildPPG) without requiring field-specific pretraining data, as previously thought necessary~\cite{saha2025pulse}. However, gains are less consistent for blood pressure prediction, where our model underperforms in one of four settings and remains close to baseline models in the other settings, which suggests that hemodynamic estimation may require richer multimodal supervision. Overall, results indicate that high-quality multimodal supervision can serve as an effective alternative to large-scale field-specific datasets for learning robust PPG representations.

A key limitation in our setup, that may help explain the lower performance on blood pressure regression, is that pretraining on MIMIC-III primarily reflects ICU populations and may not generalize to ambulatory settings. Additionally, we use only a single session per subject due to computational constraints, prioritizing subject-level diversity over within-subject temporal variability. Although selecting the longest available session for each subject partially mitigates this, future work should incorporate more heterogeneous datasets, including hospital, ambulatory, and wearable recordings, and explore scaling to larger foundation models and pretraining corpora.

Demographic variation is another important axis of generalization. We analyze performance across skin tones using the Fitzpatrick scale (Figure~\ref{fig:demographics}), but this does not fully capture skin tone diversity or potential biases in PPG signals~\citep{ware2020racial,colvonen2021response}. While these results reveal disparities, we report them for transparency and to support responsible deployment. Addressing them likely requires more diverse demographic representation in pretraining data, improved fairness evaluation, and fine-tuning on broader datasets. Furthermore, scaling to more heterogeneous populations may also help reduce bias during representation learning.

Within our multimodal framework, we use heart and respiratory metrics from biosignals as contrastive targets, though both the range of signals and metrics can be expanded within our framework. The current 10-second window, which matches the PaPaGei models, limits reliable HRV feature extraction~\citep{schroeder2004repeatability}.  Extending the PPG window to 1–2 minute windows would enable richer HRV features for pretraining, and a larger context window that can help improve embedding quality. Additional signals such as blood pressure could further enhance supervision~\cite{elgendi2019use}, especially given their availability in ICU datasets and relevance to wearable monitoring. Blood pressure regression is also a downstream task that our model currently performs closest to baselines on, further underlining its importance as a potential addition as a multimodal pretraining target. On the other hand, we find that ECG-based supervision alone already yields strong performance, suggesting diminishing returns from adding modalities in our pretraining framework. This highlights a cost–performance trade-off, as each additional biosignal increases acquisition, synchronization, and preprocessing complexity without guaranteed proportional gains, based on our results. Since our targets are aggregated physiological features (e.g., HR, RMSSD) rather than raw event-level annotations, the framework is also relatively robust to fine-grained preprocessing differences, though different reasonable pipelines may still affect supervision quality. These preprocessing pipelines are an important direction to explore in future work.

The within-subject setting we propose exhibits high variance due to limited subjects and inter-subject heterogeneity, making statistical testing underpowered despite consistent improvements across tasks. This highlights the importance of evaluating in a within-subject setting, as results may vary substantially depending on the specific subject under consideration. Future work should therefore expand evaluation to a larger and more diverse set of subjects and increase the number of evaluation folds to increase the statistical power of the analyses. More comprehensive statistical evaluation would also strengthen robustness claims in this regime. Moreover, given the heterogeneity across subjects, it may be beneficial to explicitly model subject-specific characteristics via subject-adaptive methods, particularly in low-data personalization settings.

Overall, this work shows that multimodal physiological supervision significantly improves PPG foundation models by providing physiologically grounded contrastive targets from ECG and respiratory signals. This enables more generalizable representations than PPG-only or morphology-based approaches, achieving strong performance across cross-domain settings without requiring field-specific datasets. The results highlight multimodal supervision as a scalable alternative to dataset scaling, while pointing to future directions in more diverse pretraining data, improved subject-adaptive learning, and more rigorous evaluation under distribution shift and fairness constraints.

\clearpage
\section*{Impact statement}
Non-invasive health monitoring can revolutionize the healthcare system. It is important to ensure that groups of people can equally benefit from non-invasive health monitoring. To verify biases in systolic blood pressure regression, we perform a demographic analysis to understand what demographics our model is biased towards. Although this is a step in the right direction, it is important to develop methods that can counteract any biases and more thoroughly verify what biases exist in health foundation models. Further expansions of bias analyses across more demographics, and for more tasks is thus important, and the development of datasets that allow foundation models to be tested exhaustively before deployment is essential. The deployment of wearables for health monitoring is also accompanied by ethical and legal implications that must be addressed~\citep{capulli2025ethical}. Finally, it is important that any health monitoring data from wearables is used \textbf{in alignment with a user's preferences}. We are committed to protecting participant and/or user privacy and welfare, and to ensuring scientific validity.


\bibliography{arppg}
\bibliographystyle{icml2026}

\newpage
\appendix
\onecolumn
\section{ECG and RESP pre-processing}
\label{app:multimodal-preprocessing}
We identify sessions containing more than one hour of continuous data across all three modalities: ECG, RESP, and PPG. The ECG and RESP signals are filtered using NeuroKit~\citep{Makowski2021neurokit}, and then used to detect peaks: R-peaks in ECG and respiratory peaks/troughs in RESP. Signal regions without valid peaks are trimmed, and the remaining data is segmented into non-overlapping 10-second windows. From the RESP signal, we compute RR, RA, and RVT, and average them within each window. From the ECG signal, we extract HR and RMSSD. Any NaN values (i.e. in case a heart rate under 30 was observed) were linearly interpolated. Afterwards, all metrics are filtered using a low-pass filter with a cut-off frequency of 0.001Hz (since the metrics are sampled at 0.1Hz after averaging). The PPG signal is used in its raw, unfiltered form. During pretraining we map all of the metrics within the [0, 1] range to ensure that all metrics equally contribute to the distance computation. We decide lower and upper bounds for the range based on known physiological ranges, and by computing the lower and upper 4 standard deviations away from the mean across all metrics in the pretraining dataset. We land on the following ranges, which we use to map all metrics between [0, 1]: HR [30, 210], RMSSD [10, 200], RA [8, 60], RR [0, 2], and RVT [0, 0.88]. Any values outside of this range are clipped to be within the range. Lastly, during pretraining we select one session for each subject for pretraining, the names of the session files that correspond to files in the \href{https://physionet.org/files/mimic3wdb-matched/1.0/}{MIMIC Database} can be found in the \href{https://figshare.com/articles/online_resource/Supplementary_material_for_the_paper_A_robust_PPG_foundation_model_using_multimodal_physiological_supervision_/32478918?file=65050821}{Supplementary Materials}.

\section{Downstream dataset information}
\label{app:downstream-datasets}
\paragraph{WESAD} The Wearable Stress and Affect Detection (\href{https://archive.ics.uci.edu/dataset/465/wesad+wearable+stress+and+affect+detection}{WESAD}) dataset~\citep{schmidt2018introducing} contains 15 subjects recorded in a lab setting. Although the dataset records data from a variety of physiological sensors, we only select the PPG data, which is recorded with a 64Hz sensor. In terms of PPG preprocessing we follow~\citep{xu2023rebar}, whose preprocessing code is available on \href{https://github.com/maxxu05/rebar/blob/main/data/process/ppg_processdata.py}{GitHub}. We adapt the preprocessing code to obtain 10s non-overlapping segments, and we use \href{https://github.com/Nokia-Bell-Labs/papagei-foundation-model/blob/main/linearprobing/utils.py#L233}{PaPaGei's} \texttt{resample\_batch\_signal} function to resample the segments to 125Hz to match the pretraining dataset. WESAD contains 4 classes: segments that consist of a baseline recording for each subject, segments where stress is induced using a Trier Social Stress Test (TSST)~\citep{kirschbaum1993trier}, segments where amusement is induced using a set of eleven funny video clips, and segments where subjects follow guided meditation. In case of our affect prediction task, we classify between each of these four classes. In case of the stress classification type, we classify between segments where stress is induced versus all other segments.

\paragraph{PPG-DaLiA} To better understand how well heart rate can be extracted from PPG during a wide range of activities under real-life conditions,~\citep{reiss2019deep} introduced the \href{https://archive.ics.uci.edu/dataset/495/ppg+DaLiA}{PPG-DaLiA dataset}, which records PPG at 64Hz. The dataset contains data from 15 subjects that perform eight different activities: (1) Sitting still for 10 minutes (2) Ascending/descending stairs for 5 min (3) Table soccer for 5 min (4) Cycling for 8 min (5) Driving a car for 15 min (6) Having a lunch break for 30 min (7) Walking for 10 min (8) Working for 20 min. During these daily activities, both a subject's PPG and ECG signals are recorded. The ECG signal is used as the ground truth for each 8 second window, with 2 second overlap between the windows. For the heart regression task, we use this label (\texttt{'label'}) because it is provided by the dataset and has been verified and preprocessed. Although the segment window is smaller than the pretraining dataset, both our and the PaPaGei backbone architecture can easily deal with slightly shorter segments because both use global averaging. For each subject, we first filter the PPG data with \href{https://github.com/Nokia-Bell-Labs/papagei-foundation-model/blob/main/preprocessing/ppg.py#L41}{PapaGei's} \texttt{preprocess\_one\_ppg\_signal}, segment the data into 8 second windows with a 2 second overlap to match the target labels, and then use \href{https://github.com/Nokia-Bell-Labs/papagei-foundation-model/blob/main/linearprobing/utils.py#L233}{PaPaGei's} \texttt{resample\_batch\_signal} function to resample the segments to 125Hz to match the pretraining dataset. The segments are z-scored for each subject. For the daily activities classification task, we use 10 second non-overlapping segments. The labels are sampled at 4Hz (\texttt{'activity'}), and we assign a label to a specific 10 second window if 75\% or more of the window contains that specific activity. If there is no consensus on the window, we discard it. We first filter the PPG data with \href{https://github.com/Nokia-Bell-Labs/papagei-foundation-model/blob/main/preprocessing/ppg.py#L41}{PapaGei's} \texttt{preprocess\_one\_ppg\_signal}, segment the data into 10 seconds non-overlapping segments to match the target labels, and then use \href{https://github.com/Nokia-Bell-Labs/papagei-foundation-model/blob/main/linearprobing/utils.py#L233}{PaPaGei's} \texttt{resample\_batch\_signal} function to resample the segments to 125Hz to match the pretraining dataset.

\paragraph{EEVR} The Emotion Elicitation in Virtual Reality
(\href{https://melangelabiiitd.github.io/EEVR/}{EEVR}) dataset~\citep{singh2024eevr} measures PPG data at 125Hz while 37 subjects are wearing a virtual reality (VR) headset. The study consists of baseline dataset collection, a VR familiarity task, and then a set of VR stimuli with post-exposure questionnaires. To evoke specific levels of arousal and valence, the authors use annotated 360$^\circ$ videos from a public database~\citep{li2017public}, and select videos based on four emotional quadrants of the Russell circumplex of affect~\citep{russell1980circumplex}. The circumplex contains two dimensions, valence and arousal, and the videos can thus be organized into high valence and low valence or high arousal and low arousal. The authors provide a csv file called \texttt{Raw\_PPG.csv}. We use the \texttt{Participant\_ID} column to separate data into specific subjects, and \texttt{Label\_no\_index} to separate each subject's session into a specific video with a high/low arousal and high/low valence label. Each video's corresponding PPG data is first filtered with \href{https://github.com/Nokia-Bell-Labs/papagei-foundation-model/blob/main/preprocessing/ppg.py#L41}{PapaGei's} \texttt{preprocess\_one\_ppg\_signal}, and data is segmented into 10 second, 5 second overlapping windows. Each segment is labeled separately for arousal and valence, and since the PPG sampling rate matches that of our pretraining dataset, we do not resample the data.

\paragraph{PPG-BP} To better understand how PPG can be used to understand and predict cardiovascular disease,~\citep{liang2018new} released the PPG blood pressure dataset (
\href{https://figshare.com/articles/dataset/PPG-BP_Database_zip/5459299}{PPG-BP}), with PPG sampled at 1000Hz. There are three PPG recordings for each subject that last around 2 second each, and 219 subjects in total. We noticed some issues with resampling the data, so we decided to linearly interpolate the data instead. Using \texttt{np.interp}~\citep{harris2020array}, we interpolate each segment's frequency from 1000Hz down to 125Hz to match the pretraining dataset frequency. Given that both our and PaPaGei's architecture use global averaging, both architectures can handle a variety of input sizes, so we didn't pad the input data, but before interpolation we did ensure the data was not longer than 2.1 seconds. Each subject has a recorded systolic blood pressure, diastolic blood pressure, and hypertension label. We use the same label for each of the three segments for a subject. After downsampling, we filter each segment with \href{https://github.com/Nokia-Bell-Labs/papagei-foundation-model/blob/main/preprocessing/ppg.py#L41}{PapaGei's} \texttt{preprocess\_one\_ppg\_signal}, and z-score each segment.

\paragraph{VitalVideos} As an additional evaluation of blood pressure, we also evaluate our model on the VV-Small subset of the \href{https://vitalvideos.org}{VitalVideos database}. The demographics of this dataset are outlined on Page 6 of~\citep{toye2023vital}. The dataset contains systolic and diastolic blood pressure measurements for 100 subjects, and PPG sampled at 55-60Hz. Given that the sampling rate varies throughout recording, we interpolate the data to 125Hz, in order to match the sampling rate of the pretraining dataset, with \texttt{np.interp}~\citep{harris2020array} based on the provided sample timings. We also record the age, Fitzpatrick scale, and sex of each participant to perform our demographic analysis (See Figure~\ref{fig:demographics}). After interpolating, the PPG data for each subject is filtered with \href{https://github.com/Nokia-Bell-Labs/papagei-foundation-model/blob/main/preprocessing/ppg.py#L41}{PapaGei's} \texttt{preprocess\_one\_ppg\_signal}, and data is segmented into non-overlapping 10 second windows. The PPG data is then z-scored for each subject. 

\paragraph{WildPPG} To better understand how different placements of PPG sensors, different types (wavelengths) of PPG sensors, and daily activities impact heart rate estimation~\citep{meier2024wildppg} released the \href{https://siplab.org/projects/WildPPG}{WildPPG database}. The dataset records data from 16 subjects, and each PPG sensor records at 128 Hz. The ground truth estimate of the heart rate is estimated with an ECG trace recorded from each subject's sternum. The dataset contains data for three types of PPG sensors: green, red, and infrared (IR), and four types of locations: wrist, head, ankle, and the sternum (chest in our manuscript). We follow the code provided by the authors on \href{https://github.com/eth-siplab/WildPPG/blob/main/src/supervised_baselines/data_preprocessing.py}{GitHub}, but adapt the code in the following ways. We ensure that the ground truth heart rate from the ECG trace is estimated in 10s non-overlapping windows. Moreover, for each sensor, location, and subject, we filter the PPG data with \href{https://github.com/Nokia-Bell-Labs/papagei-foundation-model/blob/main/preprocessing/ppg.py#L41}{PapaGei's} \texttt{preprocess\_one\_ppg\_signal}, segment the PPG data into non-overlapping 10s windows, and resample the segments from 128Hz to 125Hz to match the pretraining dataset with \href{https://github.com/Nokia-Bell-Labs/papagei-foundation-model/blob/main/linearprobing/utils.py#L233}{PaPaGei's} \texttt{resample\_batch\_signal}. In case it is necessary, we trim the ground-truth heart rate segments based on the number of PPG windows. Then, we remove any segments where the ground-truth heart rate is zero (generally indicates that the heart rate could not be estimated), and then we z-score the PPG data for each sensor, location, and subject.

\section{Hyperparameters and experimental settings}
\label{app:hyperparameters}
\begin{wraptable}[16]{r}{0.3\textwidth}
\centering
\caption{The 1D ResNet-26 hyperparameters} 
\begin{tabular}{l|l}
in\_channels & 128 \\
kernel\_size & 11 \\
stride & 2 \\
groups & 1 \\
n\_block & 12 \\
n\_classes & 512 \\
downsample\_gap & 2 \\
increasefilter\_gap & 4 \\
use\_bn & True \\
use\_do & True \\
verbose & False 
\end{tabular} 
\label{tab:hyperparameters} 
\end{wraptable}

\textbf{Architecture.} Our model's architecture is implemented in PyTorch~\citep{paszke2019pytorch}, and consists of two main parts. First, the input to our model is a $(2 \times \text{batch\_size}, 1, 1250)$ tensor. We use 256 as the batch size during pretraining for all models. The reason we have twice as many segments along the batch dimension is because we sample two random augmentations, as described in Section~\ref{sec:methods}. This tensor first passes through an InstanceNorm1d layer, and then the 1D ResNet-26 architecture, as described in Section~\ref{sec:methods}. The hyperparameters for the ResNet are shown in Table~\ref{tab:hyperparameters}. The output embedding for our model is thus a $(2 \times \text{batch\_size}, 512)$ tensor.

\textbf{Checkpoint selection.} During pretraining we save checkpoints for the backbone every 5000 steps. To select the final checkpoint that we use for comparisons, we evaluate each checkpoint on every single downstream cross-subject task, except the WildPPG tasks. To ensure there is no data leakage, we use each model's training and validation set score during the hyperparameter selection process for the linear probe. Moreover, we z-score the performance for all the tasks independently across the checkpoints, average the task scores for each checkpoint, and select the checkpoint with the highest average z-score. The reason we do not use the WildPPG data is to ensure atleast one dataset is not used for checkpoint selection.

\textbf{Linear probing K-folds.} There are generally two types of datasets. The following datasets where each subject has a label. PPG-BP: average heart rate (Avg-HR), systolic blood pressure (Sys-BP), diastolic blood pressure (Dia-BP), and hypertension, and VitalVideos: systolic blood pressure (Sys-BP VV) and diastolic blood pressure (Dia-BP VV). The other type of dataset are datasets with labels that vary for each subject over time. WESAD: stress and affect, PPG-DaLiA: activities and heart rate, EEVR: arousal and valence, WildPPG: heart rate. For across subject linear probing, datasets where each subject has a label are stratified when computing 5 folds. Specifically, for regression tasks, values are binned into 10 bins, using an ordinal encoding, and based on quantiles in the dataset using \texttt{KBinsDiscretizer}. For classification tasks, no additional binning is necessary. The 5 splits are then obtained using \texttt{StratifiedKFold} with \texttt{random\_state=42} and shuffling on. The training indices for each fold are split into training and validation indices with a training size of 0.75, \texttt{random\_state=42}, shuffling on, and by stratifying the targets. For datasets where labels vary over time, we obtain 5 splits with \texttt{KFold}, \texttt{random\_state=42}, and shuffling on. All names align with \texttt{scikit-learn}'s API~\citep{scikit-learn}. For within-subject linear probing, we first randomly shuffle all subjects with \texttt{np.random.default\_rng(42)}, and select the subject that corresponds to the fold index. For regression tasks, we follow the same binning process described above to create a stratified test set. The test set is created using \texttt{train\_test\_split}, stratification, 0.2 as the test size, \texttt{random\_state=42}, and shuffling on. The leftover data samples are then split into a stratified training and validation set with \texttt{train\_test\_split}, 0.8 as the training size, \texttt{random\_state=42}, and shuffling on.

\textbf{Linear probing hyperparameters.} The search space includes \(\alpha \in \{0.1, 1, 10, 100, 1000\}\) and \texttt{solver} \(\in\) \{\texttt{auto}, \texttt{cholesky}, \texttt{sparse\_cg}\}. For regression tasks, we employ ridge regression with hyperparameters \(C \in \{0.0, 0.1, 1, 10, 100\}\) and \texttt{max\_iter} fixed to 10,000. The naming of these hyperparameters is aligned with the \texttt{scikit-learn} API~\citep{scikit-learn}.

\clearpage
\section{Heart rate estimation with NeuroKit ablation}
\label{app:neurokit}
Heart rate estimation is often done using automated tools, but in cases where PPG segments are quite noisy, they may fail. In Table~\ref{tab:ablation-neurokit} we compare the best-performing PaPaGei model in terms of heart rate estimation (PaPaGei-P) with NeuroKit's automatic heart rate estimation, and our proposed model. All models take 10s of PPG segments as input, and for NeuroKit~\citep{Makowski2021neurokit} we use \texttt{ppg\_process}'s \texttt{PPG\_Rate} output. If no heart rate was detected or not enough peaks were present for NeuroKit, the heart rate was set to 0. Afterwards, we perform the linear probing procedure to account for small linear errors in the NeuroKit model, and to make the procedure as similar to the results reported for PaPaGei-P and our model. Although the example is a little manufactured given that NeuroKit is often used to estimate heart rate for longer segments of PPG data, our experiment provides a one-to-one comparison for real-time 10s window heart rate estimation. Moreover, Table~\ref{tab:ablation-neurokit} shows that our model outperforms both models, and that PaPaGei outperforms NeuroKit. In some cases, PaPaGei-P and NeuroKit's performances are closely matched, e.g. for HR (DaLiA) and HR-Green.
\begin{table*}[ht] 
\centering
\small
\setlength{\tabcolsep}{3.5pt}       
\renewcommand{\arraystretch}{1.05}  

\caption{\textbf{NeuroKit ablation, NeuroKit heart rate estimation vs PaPaGei-P and our model}. Results for both methods are averaged across $5$ test folds, and standard deviations can be found in Appendix~\ref{app:standard-deviations}. For the classification tasks, higher is better, and evaluation metrics are macro F-1 score (MF1), accuracy (ACC), and the area under the receiver operating characteristic (AUC). For regression tasks lower is better, and we use mean average error (MAE), mean squared error (MSE), and mean average percentage error (MAPE).} 
\begin{tabular}{l|ggg|xxx|ggg|zzz|}
 & \multicolumn{3}{c}{\cellcolor{Gray}\textbf{NeuroKit}} & \multicolumn{3}{c}{\cellcolor{PPGPTable}\textbf{PaPaGei-P}} & \multicolumn{3}{c}{\cellcolor{Gray}\textbf{PulsePPG}} & \multicolumn{3}{c}{\cellcolor{OURTable}\textbf{Ours}} \\ 
\hline 
Reg ($\downarrow$) & \textbf{MAE} & \textbf{MSE} & \textbf{MAPE} & \textbf{MAE} & \textbf{MSE} & \textbf{MAPE} & \textbf{MAE} & \textbf{MSE} & \textbf{MAPE} & \textbf{MAE} & \textbf{MSE} & \textbf{MAPE} \\ 
\hline 
HR (Dalia) &  12.9 &  318 &  0.14 &  12.7 &  303 &  0.14 &  9.03 &  170 &  0.1 &  7.78 &  143 &  0.09 \\ 
Avg-HR &  8.67 &  116 &  0.12 &  4.76 &  40.0 &  0.07 &  4.17 &  30.9 &  0.06 &  3.8 &  26.3 &  0.05 \\ 
HR-Green &  12.1 &  256 &  0.16 &  12.2 &  266 &  0.17 &  9.47 &  188 &  0.13 &  7.61 &  149 &  0.1 \\ 
HR-Infrared &  13.1 &  289 &  0.18 &  12.6 &  273 &  0.17 &  10.8 &  221 &  0.15 &  9.82 &  206 &  0.14 \\ 
HR-Red &  13.1 &  291 &  0.18 &  12.7 &  279 &  0.17 &  12.1 &  261 &  0.17 &  10.8 &  231 &  0.15 \\ 
\hline 
\rowcolor{white} 
Avg & 11.9  & 254  & 0.16  & 11.0  & 232  & 0.14  & 9.1  & 174  & 0.12  & \textbf{7.95}  & \textbf{151}  & \textbf{0.11}  \\ 
\hline 
\end{tabular} 
\label{tab:ablation-neurokit}
\end{table*}

\section{Expanded baseline results}
\label{app:baselines}
Table~\ref{tab:linear-probing-baselines} show the comparisons with BYOL and SimCLR. These models are trained on the same data as our own model. We find that our model in general outperforms both.
\begin{table*}[hb] 
\small
\centering
\caption{\textbf{Additional baseline results}. Results for each method are averaged across $5$ test folds, and standard deviations can be found in Appendix~\ref{app:standard-deviations}. For the classification tasks, higher is better, and evaluation metrics are macro F-1 score (MF1), accuracy (ACC), and the area under the receiver operating characteristic (AUC). For regression tasks lower is better, and we use mean average error (MAE), mean squared error (MSE), and mean average percentage error (MAPE).} 
\begin{tabular}{l|ggg|ggg|zzz|}
 & \multicolumn{3}{c}{\cellcolor{Gray}\textbf{BYOL}} & \multicolumn{3}{c}{\cellcolor{Gray}\textbf{SimCLR}} & \multicolumn{3}{c}{\cellcolor{OURTable}\textbf{Ours}} \\ 
\hline 
Clf ($\uparrow$) & \textbf{MF1} & \textbf{ACC} & \textbf{AUC} & \textbf{MF1} & \textbf{ACC} & \textbf{AUC} & \textbf{MF1} & \textbf{ACC} & \textbf{AUC} \\ 
\hline 
Stress &  0.81 &  0.87 &  0.92 &  0.83 &  0.88 &  0.94 &  0.83 &  0.88 &  0.94 \\ 
Affect &  0.49 &  0.56 &  0.75 &  0.5 &  0.59 &  0.77 &  0.52 &  0.6 &  0.78 \\ 
Activities &  0.33 &  0.38 &  0.8 &  0.32 &  0.38 &  0.8 &  0.36 &  0.41 &  0.82 \\ 
Arousal &  0.51 &  0.57 &  0.55 &  0.52 &  0.56 &  0.56 &  0.53 &  0.57 &  0.57 \\ 
Valence &  0.52 &  0.61 &  0.58 &  0.53 &  0.61 &  0.58 &  0.53 &  0.62 &  0.58 \\ 
Hypertension &  0.69 &  0.73 &  0.73 &  0.68 &  0.72 &  0.74 &  0.71 &  0.75 &  0.77 \\ 
\hline 
\rowcolor{white} 
Avg & 0.56  & 0.62  & 0.72  & 0.56  & 0.62  & 0.73  & \textbf{0.58}  & \textbf{0.64}  & \textbf{0.74}  \\ 
\hline 
Reg ($\downarrow$) & \textbf{MAE} & \textbf{MSE} & \textbf{MAPE} & \textbf{MAE} & \textbf{MSE} & \textbf{MAPE} & \textbf{MAE} & \textbf{MSE} & \textbf{MAPE} \\ 
\hline 
HR (Dalia) &  9.08 &  176 &  0.1 &  9.49 &  186 &  0.11 &  7.78 &  143 &  0.09 \\ 
Avg-HR &  3.98 &  29.3 &  0.05 &  4.4 &  33.6 &  0.06 &  3.8 &  26.3 &  0.05 \\ 
Sys-BP &  13.9 &  314 &  0.11 &  13.6 &  308 &  0.11 &  13.2 &  281 &  0.11 \\ 
Dia-BP &  8.26 &  110 &  0.12 &  8.31 &  113 &  0.12 &  8.16 &  109 &  0.12 \\ 
Sys-BP (VV) &  16.8 &  492 &  0.13 &  16.0 &  458 &  0.12 &  15.9 &  451 &  0.12 \\ 
Dia-BP (VV) &  8.44 &  125 &  0.1 &  8.17 &  122 &  0.1 &  8.04 &  123 &  0.1 \\ 
HR-Green &  8.67 &  171 &  0.12 &  8.8 &  174 &  0.12 &  7.61 &  149 &  0.1 \\ 
HR-Infrared &  10.7 &  228 &  0.15 &  10.8 &  230 &  0.15 &  9.82 &  206 &  0.14 \\ 
HR-Red &  11.4 &  248 &  0.16 &  11.5 &  250 &  0.16 &  10.8 &  231 &  0.15 \\ 
\hline 
\rowcolor{white} 
Avg & 10.1  & 210  & 0.12  & 10.1  & 208  & 0.12  & \textbf{9.45}  & \textbf{191}  & \textbf{0.11}  \\ 
\hline 
\end{tabular} 
\label{tab:linear-probing-baselines} 
\end{table*}

\section{Contrastive learning ablation}
\label{app:regression-ablation}
Table~\ref{tab:linear-probing-l2} shows the comparison between using an L2 loss during pretraining to directly predict the metrics we extract, compared to using the contrastive loss we discuss in Section~\ref{sec:multimodal-supervision}.
\renewcommand{\arraystretch}{0.8}
\begin{table}[ht] 
\centering
\small
\caption{\textbf{Our model with L2 vs the contrastive loss}. Results for each method are averaged across $5$ test folds, and standard deviations can be found in Appendix~\ref{app:standard-deviations}. For the classification tasks, higher is better, and evaluation metrics are macro F-1 score (MF1), accuracy (ACC), and the area under the receiver operating characteristic (AUC). For regression tasks lower is better, and we use mean average error (MAE), mean squared error (MSE), and mean average percentage error (MAPE).} 
\begin{tabular}{l|ggg|zzz|}
 & \multicolumn{3}{c}{\cellcolor{Gray}\textbf{Ours (L2 loss)}} & \multicolumn{3}{c}{\cellcolor{OURTable}\textbf{Ours (new)}} \\ 
\hline 
Clf ($\uparrow$) & \textbf{MF1} & \textbf{ACC} & \textbf{AUC} & \textbf{MF1} & \textbf{ACC} & \textbf{AUC} \\ 
\hline 
Stress &  0.81 &  0.86 &  0.92 &  0.83 &  0.88 &  0.94 \\ 
Affect &  0.5 &  0.58 &  0.76 &  0.52 &  0.6 &  0.78 \\ 
Activities &  0.31 &  0.37 &  0.79 &  0.36 &  0.41 &  0.82 \\ 
Arousal &  0.51 &  0.56 &  0.54 &  0.53 &  0.57 &  0.57 \\ 
Valence &  0.53 &  0.61 &  0.59 &  0.53 &  0.62 &  0.58 \\ 
Hypertension &  0.69 &  0.73 &  0.73 &  0.71 &  0.75 &  0.77 \\ 
\hline 
\rowcolor{white} 
Avg & 0.56  & 0.62  & 0.72  & \textbf{0.58}  & \textbf{0.64}  & \textbf{0.74}  \\ 
\hline 
Reg ($\downarrow$) & \textbf{MAE} & \textbf{MSE} & \textbf{MAPE} & \textbf{MAE} & \textbf{MSE} & \textbf{MAPE} \\ 
\hline 
HR (Dalia) &  8.39 &  163 &  0.09 &  7.78 &  143 &  0.09 \\ 
Avg-HR &  3.84 &  26.1 &  0.05 &  3.8 &  26.3 &  0.05 \\ 
Sys-BP &  14.1 &  330 &  0.11 &  13.2 &  281 &  0.11 \\ 
Dia-BP &  8.64 &  121 &  0.12 &  8.16 &  109 &  0.12 \\ 
Sys-BP (VV) &  15.6 &  433 &  0.12 &  15.9 &  451 &  0.12 \\ 
Dia-BP (VV) &  8.37 &  126 &  0.1 &  8.04 &  123 &  0.1 \\ 
HR-Green &  8.52 &  173 &  0.12 &  7.61 &  149 &  0.1 \\ 
HR-Infrared &  10.8 &  237 &  0.15 &  9.82 &  206 &  0.14 \\ 
HR-Red &  11.6 &  255 &  0.16 &  10.8 &  231 &  0.15 \\ 
\hline 
\rowcolor{white} 
Avg & 9.98  & 207  & 0.11  & \textbf{9.45}  & \textbf{191}  & \textbf{0.11}  \\ 
\hline 
\end{tabular} 
\label{tab:linear-probing-l2} 
\end{table}

\section{Projector ablation}
\label{app:projector-ablation}
We use a projector, like the BYOL and SimCLR models, during pretraining. The following experiment ablates the use of this projector. We found gains in both the classification and regression tasks, and use this model for all experiments. The results are shown in Table~\ref{tab:linear-probing-new}.
\renewcommand{\arraystretch}{0.8}
\begin{table}[ht]
\centering
\small
\caption{\textbf{Projector ablation across subjects linear probing results}. Results for each method are averaged across $5$ test folds, and standard deviations can be found in Appendix~\ref{app:standard-deviations}. For the classification tasks, higher is better, and evaluation metrics are macro F-1 score (MF1), accuracy (ACC), and the area under the receiver operating characteristic (AUC). For regression tasks lower is better, and we use mean average error (MAE), mean squared error (MSE), and mean average percentage error (MAPE).} 
\begin{tabular}{l|ggg|zzz|}
 & \multicolumn{3}{c}{\cellcolor{Gray}\textbf{Without projector}} & \multicolumn{3}{c}{\cellcolor{OURTable}\textbf{With projector}} \\ 
\hline 
Clf ($\uparrow$) & \textbf{MF1} & \textbf{ACC} & \textbf{AUC} & \textbf{MF1} & \textbf{ACC} & \textbf{AUC} \\ 
\hline 
Stress &  0.82 &  0.87 &  0.93 &  0.83 &  0.88 &  0.94 \\ 
Affect &  0.53 &  0.62 &  0.78 &  0.52 &  0.6 &  0.78 \\ 
Activities &  0.35 &  0.41 &  0.81 &  0.36 &  0.41 &  0.82 \\ 
Arousal &  0.53 &  0.56 &  0.55 &  0.53 &  0.57 &  0.57 \\ 
Valence &  0.55 &  0.62 &  0.59 &  0.53 &  0.62 &  0.58 \\ 
Hypertension &  0.68 &  0.72 &  0.75 &  0.71 &  0.75 &  0.77 \\ 
\hline 
\rowcolor{white} 
Avg & 0.58  & 0.63  & 0.74  & \textbf{0.58}  & \textbf{0.64}  & \textbf{0.74}  \\ 
\hline 
Reg ($\downarrow$) & \textbf{MAE} & \textbf{MSE} & \textbf{MAPE} & \textbf{MAE} & \textbf{MSE} & \textbf{MAPE} \\ 
\hline 
HR (Dalia) &  7.2 &  134 &  0.08 &  7.78 &  143 &  0.09 \\ 
Avg-HR &  3.74 &  24.9 &  0.05 &  3.8 &  26.3 &  0.05 \\ 
Sys-BP &  13.5 &  299 &  0.11 &  13.2 &  281 &  0.11 \\ 
Dia-BP &  8.38 &  115 &  0.12 &  8.16 &  109 &  0.12 \\ 
Sys-BP (VV) &  16.6 &  473 &  0.12 &  15.9 &  451 &  0.12 \\ 
Dia-BP (VV) &  9.0 &  145 &  0.11 &  8.04 &  123 &  0.1 \\ 
HR-Green &  7.28 &  144 &  0.1 &  7.61 &  149 &  0.1 \\ 
HR-Infrared &  9.57 &  201 &  0.13 &  9.82 &  206 &  0.14 \\ 
HR-Red &  10.6 &  227 &  0.15 &  10.8 &  231 &  0.15 \\ 
\hline 
\rowcolor{white} 
Avg & 9.54  & 196  & 0.11  & \textbf{9.45}  & \textbf{191}  & \textbf{0.11}  \\ 
\hline 
\end{tabular} 
\label{tab:linear-probing-new} 
\end{table}

\section{Pretraining ablation with the same architecture}
\label{app:ablation:pretraining}
To compare PaPaGei's unimodal pretraining loss with our multimodal pretraining loss, we adopt the PaPaGei-S pretraining objective, and match the pretraining data, and all training hyperparameters, including the architectures, between the models. The setups only differ in the pretraining objective. As detailed in Table~\ref{tab:ablation-unimodal}, our multimodal pretraining consistently and substantially outperforms unimodal pretraining across all evaluated tasks, except diastolic blood pressure regression.
\renewcommand{\arraystretch}{0.8}
\begin{table}[hb]
\centering
\small
\caption{\textbf{Unimodal vs multimodal pre-training, same architecture and data}. Results for both methods are averaged across $5$ test folds, and standard deviations can be found in Appendix~\ref{app:standard-deviations}. For the classification tasks, higher is better, and evaluation metrics are macro F-1 score (MF1), accuracy (ACC), and the area under the receiver operating characteristic (AUC). For regression tasks lower is better, and we use mean average error (MAE), mean squared error (MSE), and mean average percentage error (MAPE).} 
\begin{tabular}{l|ggg|zzz|}
 & \multicolumn{3}{c}{\cellcolor{Gray}\textbf{Unimodal pre-training}} & \multicolumn{3}{c}{\cellcolor{OURTable}\textbf{Multimodal pre-training}} \\ 
\hline 
Clf ($\uparrow$) & \textbf{MF1} & \textbf{ACC} & \textbf{AUC} & \textbf{MF1} & \textbf{ACC} & \textbf{AUC} \\ 
\hline 
Stress &  0.63 &  0.79 &  0.84 &  0.76 &  0.83 &  0.86 \\ 
Affect &  0.39 &  0.51 &  0.71 &  0.43 &  0.53 &  0.71 \\ 
Activities &  0.19 &  0.32 &  0.7 &  0.31 &  0.37 &  0.78 \\ 
Arousal &  0.38 &  0.58 &  0.53 &  0.49 &  0.55 &  0.52 \\ 
Valence &  0.39 &  0.65 &  0.53 &  0.52 &  0.61 &  0.57 \\ 
Hypertension &  0.52 &  0.64 &  0.61 &  0.67 &  0.72 &  0.72 \\ 
\hline 
\rowcolor{white} 
Avg & 0.42  & 0.58  & 0.65  & \textbf{0.53}  & \textbf{0.6}  & \textbf{0.69}  \\ 
\hline 
Reg ($\downarrow$) & \textbf{MAE} & \textbf{MSE} & \textbf{MAPE} & \textbf{MAE} & \textbf{MSE} & \textbf{MAPE} \\ 
\hline 
HR (Dalia) &  16.0 &  443 &  0.18 &  8.25 &  167 &  0.09 \\ 
Avg-HR &  7.69 &  94.6 &  0.11 &  3.69 &  24.2 &  0.05 \\ 
Sys-BP &  15.8 &  408 &  0.13 &  14.1 &  326 &  0.11 \\ 
Dia-BP &  8.58 &  120 &  0.12 &  8.73 &  120 &  0.12 \\ 
Sys-BP (VV) &  16.9 &  505 &  0.13 &  16.2 &  467 &  0.12 \\ 
Dia-BP (VV) &  8.35 &  127 &  0.1 &  8.64 &  137 &  0.11 \\ 
HR-Green &  13.0 &  286 &  0.18 &  8.42 &  176 &  0.12 \\ 
HR-Infrared &  12.8 &  277 &  0.18 &  10.4 &  226 &  0.15 \\ 
HR-Red &  12.7 &  279 &  0.17 &  11.6 &  260 &  0.16 \\ 
\hline 
\rowcolor{white} 
Avg & 12.4  & 282  & 0.14  & \textbf{10.0}  & \textbf{211}  & \textbf{0.11}  \\ 
\hline 
\end{tabular} 
\label{tab:ablation-unimodal}
\end{table}

\section{Architecture ablation with the same pretraining}
\label{app:ablation:architecture}
To verify that our model's performance scales well with backbone size, we compare the PaPaGei backbone with our proposed backbone in Table~\ref{tab:ablation-architecture}, and use our proposed pretraining objective. The proposed larger backbone improves over the PaPaGei backbone on all but the average heart rate estimation, which indicates that our pretraining method likely benefits from further model size scaling.
\renewcommand{\arraystretch}{0.8}
\begin{table}[hb]
\small
\centering
\caption{\textbf{Architecture ablation, PaPaGei backbone architecture vs our proposed architecture, same pre-training}. Results for both methods are averaged across $5$ test folds, and standard deviations can be found in Appendix~\ref{app:standard-deviations}. For the classification tasks, higher is better, and evaluation metrics are macro F-1 score (MF1), accuracy (ACC), and the area under the receiver operating characteristic (AUC). For regression tasks lower is better, and we use mean average error (MAE), mean squared error (MSE), and mean average percentage error (MAPE).} 
\begin{tabular}{l|ggg|zzz|}
 & \multicolumn{3}{c}{\cellcolor{Gray}\textbf{PaPaGei Arch}} & \multicolumn{3}{c}{\cellcolor{OURTable}\textbf{Proposed Arch}} \\ 
\hline 
Clf ($\uparrow$) & \textbf{MF1} & \textbf{ACC} & \textbf{AUC} & \textbf{MF1} & \textbf{ACC} & \textbf{AUC} \\ 
\hline 
Stress &  0.76 &  0.83 &  0.86 &  0.83 &  0.88 &  0.94 \\ 
Affect &  0.43 &  0.53 &  0.71 &  0.52 &  0.6 &  0.78 \\ 
Activities &  0.31 &  0.37 &  0.78 &  0.36 &  0.41 &  0.82 \\ 
Arousal &  0.49 &  0.55 &  0.52 &  0.53 &  0.57 &  0.57 \\ 
Valence &  0.52 &  0.61 &  0.57 &  0.53 &  0.62 &  0.58 \\ 
Hypertension &  0.67 &  0.72 &  0.72 &  0.71 &  0.75 &  0.77 \\ 
\hline 
\rowcolor{white} 
Avg & 0.53  & 0.6  & 0.69  & \textbf{0.58}  & \textbf{0.64}  & \textbf{0.74}  \\ 
\hline 
Reg ($\downarrow$) & \textbf{MAE} & \textbf{MSE} & \textbf{MAPE} & \textbf{MAE} & \textbf{MSE} & \textbf{MAPE} \\ 
\hline 
HR (Dalia) &  8.25 &  167 &  0.09 &  7.78 &  143 &  0.09 \\ 
Avg-HR &  3.69 &  24.2 &  0.05 &  3.8 &  26.3 &  0.05 \\ 
Sys-BP &  14.1 &  326 &  0.11 &  13.2 &  281 &  0.11 \\ 
Dia-BP &  8.73 &  120 &  0.12 &  8.16 &  109 &  0.12 \\ 
Sys-BP (VV) &  16.2 &  467 &  0.12 &  15.9 &  451 &  0.12 \\ 
Dia-BP (VV) &  8.64 &  137 &  0.11 &  8.04 &  123 &  0.1 \\ 
HR-Green &  8.42 &  176 &  0.12 &  7.61 &  149 &  0.1 \\ 
HR-Infrared &  10.4 &  226 &  0.15 &  9.82 &  206 &  0.14 \\ 
HR-Red &  11.6 &  260 &  0.16 &  10.8 &  231 &  0.15 \\ 
\hline 
\rowcolor{white} 
Avg & 10.0  & 211  & 0.11  & \textbf{9.45}  & \textbf{191}  & \textbf{0.11}  \\ 
\hline 
\end{tabular} 
\label{tab:ablation-architecture} 
\end{table}

\section{Additional within-subject generalization results}
\label{app:within-subject}
\begin{figure}[ht]
\centering
\includegraphics[width=0.4\textwidth]{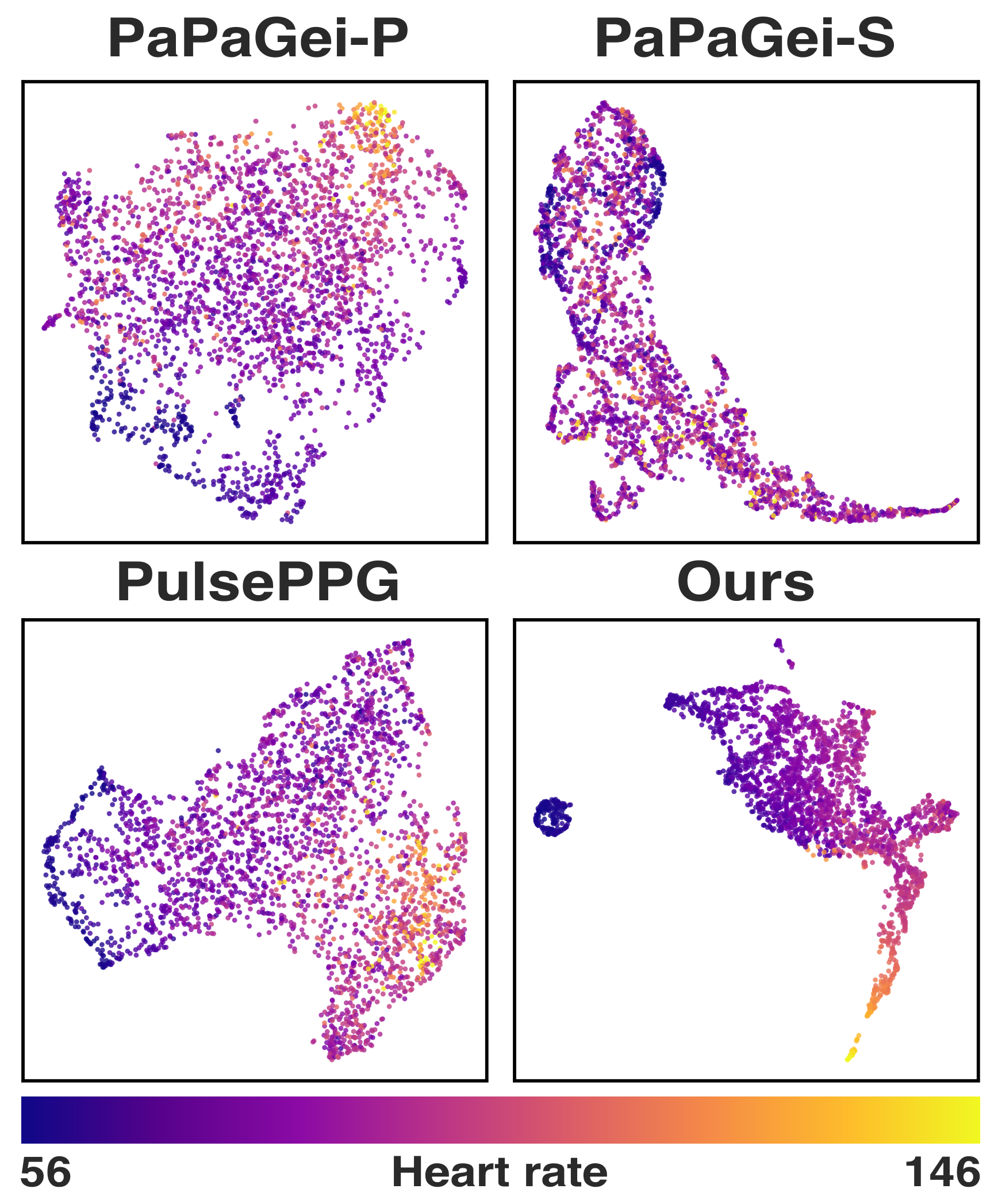} 
\caption{\textbf{UMAP plots\citep{mcinnes2018umap} colored by heart rate.} Data are from a single subject in the PPG-DaLiA dataset.
}
\label{fig:umap}
\includegraphics[width=\textwidth]{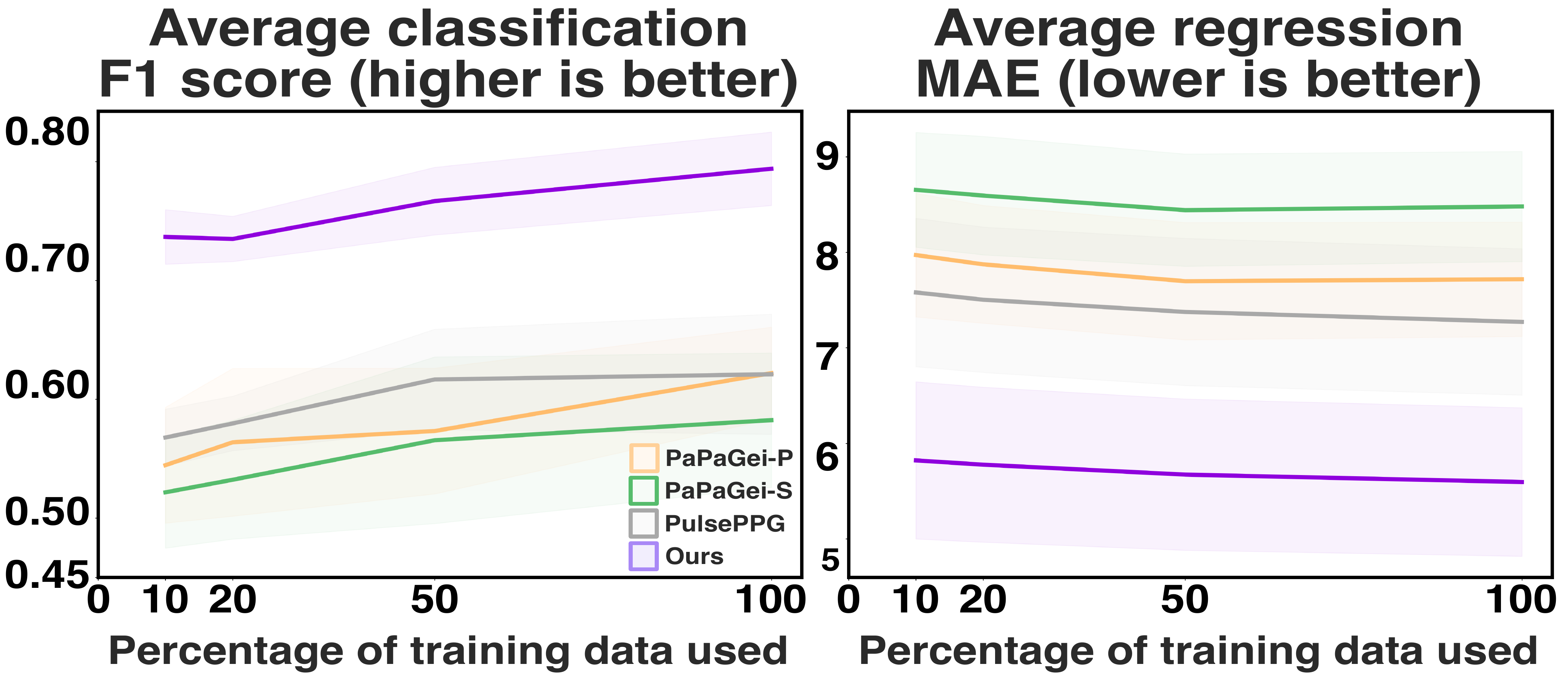} 
\caption{\textbf{Average performance across varying percentages of within-subject data.} Shaded areas represent standard deviation across folds.
}
\label{fig:subjectdata}
\end{figure}

Figure~\ref{fig:umap} shows a visualization of the embedding space for a single subject from the PPG-DaLiA dataset, highlighting the difference between our method and our replication of PaPaGei-S (labeled as 'Unimodal'). In the figure, our model's embedding space shows a clear gradient in terms of heart rate, whereas the other models do not. Moreover, since data availability for a new user may be sparse, in Figure~\ref{fig:subjectdata} we show how our model significantly outperforms the baseline models. Specifically, even on 10\% data, the average performance of our model is better than the best baseline model on 100\% of the data. Data is removed in a stratified manner from the training set.

\clearpage
\section{Standard deviations}
\label{app:standard-deviations}
Tables ~\ref{tab:linear-probing-group-sd},~\ref{tab:linear-probing-subject-sd},~\ref{tab:ablation-unimodal-sd}, and~\ref{tab:ablation-architecture-sd} report the standard deviations across 5 folds for the main tables in the text. Table~\ref{tab:linear-probing-group-sd} corresponds to Table~\ref{tab:linear-probing-group} in the main text, and Table~\ref{tab:linear-probing-subject-sd} corresponds to Table~\ref{tab:linear-probing-subject} in the main text. Tables~\ref{tab:ablation-unimodal-sd} and~\ref{tab:ablation-architecture-sd} correspond to ablation Tables~\ref{tab:ablation-unimodal} and~\ref{tab:ablation-architecture}, respectively.
\begin{table}[ht]
\small
\centering
\setlength{\tabcolsep}{3.5pt} 
\renewcommand{\arraystretch}{1.05}
\caption{\textbf{Downstream across subjects linear probing standard deviations across 5 folds}. Results for each method are averaged across $5$ test folds, and standard deviations can be found in Appendix~\ref{app:standard-deviations}. For the classification tasks, higher is better, and evaluation metrics are macro F-1 score (MF1), accuracy (ACC), and the area under the receiver operating characteristic (AUC). For regression tasks lower is better, and we use mean average error (MAE), mean squared error (MSE), and mean average percentage error (MAPE).} 
\begin{tabular}{l|xxx|yyy|ggg|zzz|}
 & \multicolumn{3}{c}{\cellcolor{PPGPTable}\textbf{PaPaGei-P}} & \multicolumn{3}{c}{\cellcolor{PPGSTable}\textbf{PaPaGei-S}} & \multicolumn{3}{c}{\cellcolor{Gray}\textbf{PulsePPG}} & \multicolumn{3}{c}{\cellcolor{OURTable}\textbf{Ours}} \\ 
\hline 
Clf ($\uparrow$) & \textbf{MF1} & \textbf{ACC} & \textbf{AUC} & \textbf{MF1} & \textbf{ACC} & \textbf{AUC} & \textbf{MF1} & \textbf{ACC} & \textbf{AUC} & \textbf{MF1} & \textbf{ACC} & \textbf{AUC} \\ 
\hline 
Stress &  0.03 &  0.02 &  0.06 &  0.04 &  0.03 &  0.05 &  0.04 &  0.03 &  0.02 &  0.04 &  0.03 &  0.02 \\ 
Affect &  0.05 &  0.04 &  0.04 &  0.01 &  0.03 &  0.03 &  0.02 &  0.04 &  0.05 &  0.05 &  0.04 &  0.05 \\ 
Activities &  0.04 &  0.03 &  0.03 &  0.03 &  0.03 &  0.03 &  0.06 &  0.05 &  0.02 &  0.04 &  0.04 &  0.03 \\ 
Arousal &  0.02 &  0.01 &  0.02 &  0.02 &  0.02 &  0.02 &  0.03 &  0.03 &  0.01 &  0.03 &  0.01 &  0.04 \\ 
Valence &  0.02 &  0.01 &  0.04 &  0.02 &  0.02 &  0.02 &  0.01 &  0.02 &  0.02 &  0.02 &  0.04 &  0.05 \\ 
Hypertension &  0.03 &  0.03 &  0.04 &  0.03 &  0.04 &  0.05 &  0.04 &  0.03 &  0.04 &  0.05 &  0.04 &  0.04 \\ 
\hline 
\rowcolor{white} 
Avg & 0.03  & 0.03  & 0.04  & 0.03  & 0.03  & 0.03  & 0.03  & 0.03  & 0.03  & 0.04  & 0.03  & 0.04  \\ 
\hline 
Reg ($\downarrow$) & \textbf{MAE} & \textbf{MSE} & \textbf{MAPE} & \textbf{MAE} & \textbf{MSE} & \textbf{MAPE} & \textbf{MAE} & \textbf{MSE} & \textbf{MAPE} & \textbf{MAE} & \textbf{MSE} & \textbf{MAPE} \\ 
\hline 
HR (Dalia) &  1.08 &  75.7 &  0.02 &  1.57 &  111 &  0.03 &  0.98 &  45.8 &  0.02 &  0.89 &  41.0 &  0.02 \\ 
Avg-HR &  0.27 &  6.04 &  0.0 &  0.35 &  3.56 &  0.0 &  0.24 &  4.41 &  0.0 &  0.23 &  3.63 &  0.0 \\ 
Sys-BP &  0.8 &  57.0 &  0.01 &  0.62 &  47.5 &  0.01 &  1.01 &  59.8 &  0.01 &  1.16 &  47.8 &  0.01 \\ 
Dia-BP &  0.34 &  6.61 &  0.01 &  0.31 &  6.44 &  0.01 &  0.35 &  8.71 &  0.01 &  0.34 &  6.18 &  0.01 \\ 
Sys-BP (VV) &  2.0 &  148 &  0.01 &  2.18 &  179 &  0.02 &  1.13 &  69.5 &  0.01 &  1.6 &  110 &  0.01 \\ 
Dia-BP (VV) &  1.1 &  56.5 &  0.01 &  1.35 &  53.7 &  0.02 &  1.29 &  60.5 &  0.02 &  1.86 &  73.0 &  0.02 \\ 
HR-Green &  1.51 &  68.1 &  0.04 &  1.28 &  54.4 &  0.03 &  1.36 &  51.9 &  0.03 &  1.3 &  50.1 &  0.03 \\ 
HR-Infrared &  2.05 &  86.1 &  0.04 &  2.17 &  89.5 &  0.05 &  1.84 &  77.8 &  0.04 &  1.98 &  85.6 &  0.04 \\ 
HR-Red &  0.79 &  29.7 &  0.01 &  0.77 &  27.4 &  0.01 &  0.82 &  30.4 &  0.01 &  0.4 &  12.2 &  0.01 \\ 
\hline 
\rowcolor{white} 
Avg & 1.1  & 59.3  & 0.02  & 1.18  & 63.7  & 0.02  & 1.0  & 45.4  & 0.02  & 1.09  & 47.8  & 0.02  \\ 
\hline 
\end{tabular} 
\label{tab:linear-probing-group-sd} 
\end{table}

\begin{table}[ht]
\small
\centering
\setlength{\tabcolsep}{3.5pt} 
\renewcommand{\arraystretch}{1.05}
\caption{\textbf{Downstream within subjects linear probing standard deviations across 5 folds}. Results for each method are averaged across $5$ test folds, and standard deviations can be found in Appendix~\ref{app:standard-deviations}. For the classification tasks, higher is better, and evaluation metrics are macro F-1 score (MF1), accuracy (ACC), and the area under the receiver operating characteristic (AUC). For regression tasks lower is better, and we use mean average error (MAE), mean squared error (MSE), and mean average percentage error (MAPE).} 
\begin{tabular}{l|xxx|yyy|ggg|zzz|}
 & \multicolumn{3}{c}{\cellcolor{PPGPTable}\textbf{PaPaGei-P}} & \multicolumn{3}{c}{\cellcolor{PPGSTable}\textbf{PaPaGei-S}} & \multicolumn{3}{c}{\cellcolor{Gray}\textbf{PulsePPG}} & \multicolumn{3}{c}{\cellcolor{OURTable}\textbf{Ours}} \\ 
\hline 
Clf ($\uparrow$) & \textbf{MF1} & \textbf{ACC} & \textbf{AUC} & \textbf{MF1} & \textbf{ACC} & \textbf{AUC} & \textbf{MF1} & \textbf{ACC} & \textbf{AUC} & \textbf{MF1} & \textbf{ACC} & \textbf{AUC} \\ 
\hline 
Stress &  0.04 &  0.03 &  0.08 &  0.1 &  0.08 &  0.12 &  0.07 &  0.07 &  0.07 &  0.08 &  0.06 &  0.04 \\ 
Affect &  0.09 &  0.08 &  0.09 &  0.13 &  0.11 &  0.08 &  0.09 &  0.08 &  0.06 &  0.08 &  0.07 &  0.03 \\ 
Activities &  0.14 &  0.13 &  0.05 &  0.14 &  0.14 &  0.09 &  0.09 &  0.1 &  0.03 &  0.09 &  0.1 &  0.03 \\ 
Arousal &  0.02 &  0.02 &  0.03 &  0.06 &  0.06 &  0.07 &  0.03 &  0.03 &  0.03 &  0.05 &  0.05 &  0.06 \\ 
Valence &  0.08 &  0.07 &  0.1 &  0.11 &  0.12 &  0.14 &  0.14 &  0.13 &  0.15 &  0.1 &  0.1 &  0.11 \\ 
\hline 
\rowcolor{white} 
Avg & 0.07  & 0.07  & 0.07  & 0.11  & 0.1  & 0.1  & 0.08  & 0.08  & 0.07  & 0.08  & 0.08  & 0.05  \\ 
\hline 
Reg ($\downarrow$) & \textbf{MAE} & \textbf{MSE} & \textbf{MAPE} & \textbf{MAE} & \textbf{MSE} & \textbf{MAPE} & \textbf{MAE} & \textbf{MSE} & \textbf{MAPE} & \textbf{MAE} & \textbf{MSE} & \textbf{MAPE} \\ 
\hline 
HR (Dalia) &  0.88 &  26.7 &  0.01 &  1.03 &  40.0 &  0.01 &  0.78 &  16.9 &  0.01 &  0.76 &  14.6 &  0.01 \\ 
HR-Green &  0.65 &  36.0 &  0.01 &  0.38 &  48.7 &  0.01 &  0.81 &  32.1 &  0.02 &  0.94 &  32.0 &  0.02 \\ 
HR-Infrared &  1.5 &  69.0 &  0.03 &  1.54 &  80.5 &  0.03 &  1.2 &  54.4 &  0.02 &  1.56 &  63.6 &  0.03 \\ 
HR-Red &  0.44 &  28.6 &  0.01 &  0.5 &  45.8 &  0.01 &  0.81 &  39.4 &  0.01 &  0.98 &  33.8 &  0.02 \\ 
\hline 
\rowcolor{white} 
Avg & 0.87  & 40.1  & 0.02  & 0.86  & 53.7  & 0.02  & 0.9  & 35.7  & 0.02  & 1.06  & 36.0  & 0.02  \\ 
\hline 
\end{tabular} 
\label{tab:linear-probing-subject-sd} 
\end{table}

\begin{table}[ht]
\centering
\small
\setlength{\tabcolsep}{3.5pt} 
\renewcommand{\arraystretch}{1.05}
\caption{\textbf{Unimodal vs multimodal pre-training standard deviations across 5 folds}. Results for both methods are averaged across $5$ test folds, and standard deviations can be found in Appendix~\ref{app:standard-deviations}. For the classification tasks, higher is better, and evaluation metrics are macro F-1 score (MF1), accuracy (ACC), and the area under the receiver operating characteristic (AUC). For regression tasks lower is better, and we use mean average error (MAE), mean squared error (MSE), and mean average percentage error (MAPE).} 
\begin{tabular}{l|ggg|zzz|}
 & \multicolumn{3}{c}{\cellcolor{Gray}\textbf{Unimodal pre-training}} & \multicolumn{3}{c}{\cellcolor{OURTable}\textbf{Multimodal pre-training}} \\ 
\hline 
Clf ($\uparrow$) & \textbf{MF1} & \textbf{ACC} & \textbf{AUC} & \textbf{MF1} & \textbf{ACC} & \textbf{AUC} \\ 
\hline 
Stress &  0.06 &  0.02 &  0.03 &  0.04 &  0.03 &  0.04 \\ 
Affect &  0.04 &  0.03 &  0.05 &  0.04 &  0.06 &  0.04 \\ 
Activities &  0.03 &  0.04 &  0.04 &  0.05 &  0.05 &  0.03 \\ 
Arousal &  0.02 &  0.02 &  0.02 &  0.01 &  0.03 &  0.02 \\ 
Valence &  0.0 &  0.0 &  0.04 &  0.03 &  0.02 &  0.02 \\ 
Hypertension &  0.07 &  0.04 &  0.04 &  0.03 &  0.02 &  0.03 \\ 
\hline 
\rowcolor{white} 
Avg & 0.04  & 0.03  & 0.04  & 0.03  & 0.03  & 0.03  \\ 
\hline 
Reg ($\downarrow$) & \textbf{MAE} & \textbf{MSE} & \textbf{MAPE} & \textbf{MAE} & \textbf{MSE} & \textbf{MAPE} \\ 
\hline 
HR (Dalia) &  2.38 &  155 &  0.03 &  1.1 &  42.5 &  0.03 \\ 
Avg-HR &  0.37 &  11.4 &  0.01 &  0.19 &  3.58 &  0.0 \\ 
Sys-BP &  0.31 &  39.2 &  0.0 &  0.77 &  49.8 &  0.01 \\ 
Dia-BP &  0.24 &  6.02 &  0.01 &  0.41 &  9.31 &  0.01 \\ 
Sys-BP (VV) &  1.16 &  105 &  0.01 &  1.22 &  87.4 &  0.01 \\ 
Dia-BP (VV) &  1.3 &  58.4 &  0.01 &  1.27 &  64.1 &  0.02 \\ 
HR-Green &  1.21 &  52.5 &  0.03 &  1.65 &  66.3 &  0.04 \\ 
HR-Infrared &  2.18 &  88.6 &  0.05 &  1.97 &  88.0 &  0.04 \\ 
HR-Red &  0.76 &  28.2 &  0.01 &  0.79 &  28.2 &  0.01 \\ 
\hline 
\rowcolor{white} 
Avg & 1.1  & 60.5  & 0.02  & 1.04  & 48.8  & 0.02  \\ 
\hline 
\end{tabular} 
\label{tab:ablation-unimodal-sd} 
\end{table}

\begin{table}[ht]
\centering
\setlength{\tabcolsep}{3.5pt} 
\renewcommand{\arraystretch}{1.05}
\caption{\textbf{Architecture ablation standard deviations across 5 folds}. Results for both methods are averaged across $5$ test folds, and standard deviations can be found in Appendix~\ref{app:standard-deviations}. For the classification tasks, higher is better, and evaluation metrics are macro F-1 score (MF1), accuracy (ACC), and the area under the receiver operating characteristic (AUC). For regression tasks lower is better, and we use mean average error (MAE), mean squared error (MSE), and mean average percentage error (MAPE).} 
\begin{tabular}{l|ggg|zzz|}
 & \multicolumn{3}{c}{\cellcolor{Gray}\textbf{PaPaGei Arch}} & \multicolumn{3}{c}{\cellcolor{OURTable}\textbf{Proposed Arch}} \\ 
\hline 
Clf ($\uparrow$) & \textbf{MF1} & \textbf{ACC} & \textbf{AUC} & \textbf{MF1} & \textbf{ACC} & \textbf{AUC} \\ 
\hline 
Stress &  0.04 &  0.03 &  0.04 &  0.04 &  0.03 &  0.02 \\ 
Affect &  0.04 &  0.06 &  0.04 &  0.05 &  0.04 &  0.05 \\ 
Activities &  0.05 &  0.05 &  0.03 &  0.04 &  0.04 &  0.03 \\ 
Arousal &  0.01 &  0.03 &  0.02 &  0.03 &  0.01 &  0.04 \\ 
Valence &  0.03 &  0.02 &  0.02 &  0.02 &  0.04 &  0.05 \\ 
Hypertension &  0.03 &  0.02 &  0.03 &  0.05 &  0.04 &  0.04 \\ 
\hline 
\rowcolor{white} 
Avg & 0.03  & 0.03  & 0.03  & 0.04  & 0.03  & 0.04  \\ 
\hline 
Reg ($\downarrow$) & \textbf{MAE} & \textbf{MSE} & \textbf{MAPE} & \textbf{MAE} & \textbf{MSE} & \textbf{MAPE} \\ 
\hline 
HR (Dalia) &  1.1 &  42.5 &  0.03 &  0.89 &  41.0 &  0.02 \\ 
Avg-HR &  0.19 &  3.58 &  0.0 &  0.23 &  3.63 &  0.0 \\ 
Sys-BP &  0.77 &  49.8 &  0.01 &  1.16 &  47.8 &  0.01 \\ 
Dia-BP &  0.41 &  9.31 &  0.01 &  0.34 &  6.18 &  0.01 \\ 
Sys-BP (VV) &  1.22 &  87.4 &  0.01 &  1.6 &  110 &  0.01 \\ 
Dia-BP (VV) &  1.27 &  64.1 &  0.02 &  1.86 &  73.0 &  0.02 \\ 
HR-Green &  1.65 &  66.3 &  0.04 &  1.3 &  50.1 &  0.03 \\ 
HR-Infrared &  1.97 &  88.0 &  0.04 &  1.98 &  85.6 &  0.04 \\ 
HR-Red &  0.79 &  28.2 &  0.01 &  0.4 &  12.2 &  0.01 \\ 
\hline 
\rowcolor{white} 
Avg & 1.04  & 48.8  & 0.02  & 1.09  & 47.8  & 0.02  \\ 
\hline 
\end{tabular} 
\label{tab:ablation-architecture-sd} 
\end{table}

\clearpage
\section{General timeseries foundation model comparisons}
Our model achieves consistent improvements over general-purpose timeseries foundation models Chronos and MOMENT across both classification and regression tasks under across-subject linear probing. The only task that Chronos and MOMENT outperform our model at is systolic blood pressure regression on the VitalVideos dataset, which is a task that other baselines also outperform our model at. Another notable result is that MOMENT comes closest to our model's performance relative to other baselines for Avg-HR prediction, even though it is not a PPG-specific foundation model. Lastly, Chronos performs on par with our model in terms of F-1 score and AUC for valence prediction, but does worse in terms of accuracy. We hence conclude that although Chronos' performance is close to ours, it still underperforms our model in terms of valence prediction. Overall, these results are consistent with results for other baseline models, namely that our model generally outperforms across most downstream tasks, except systolic blood pressure prediction on the VitalVideos dataset. It is also important to note that these general time-series foundation models contain substantially more parameters than our model.

\label{app:foundation-model}
\begin{table}[ht] 
\caption{\textbf{Downstream across subjects linear probing results for general timeseries foundation models}. Results for each method are averaged across $5$ test folds, and standard deviations can be found in Appendix~\ref{app:standard-deviations}. For the classification tasks, higher is better, and evaluation metrics are macro F-1 score (MF1), accuracy (ACC), and the area under the receiver operating characteristic (AUC). For regression tasks lower is better, and we use mean average error (MAE), mean squared error (MSE), and mean average percentage error (MAPE).} 
\begin{tabular}{l|ggg|ggg|ggg|zzz|}
 & \multicolumn{3}{c}{\cellcolor{Gray}\textbf{PulsePPG}} & \multicolumn{3}{c}{\cellcolor{Gray}\textbf{Chronos}} & \multicolumn{3}{c}{\cellcolor{Gray}\textbf{MOMENT}} & \multicolumn{3}{c}{\cellcolor{OURTable}\textbf{Ours}} \\ 
\hline 
Clf ($\uparrow$) & \textbf{MF1} & \textbf{ACC} & \textbf{AUC} & \textbf{MF1} & \textbf{ACC} & \textbf{AUC} & \textbf{MF1} & \textbf{ACC} & \textbf{AUC} & \textbf{MF1} & \textbf{ACC} & \textbf{AUC} \\ 
\hline 
Stress &  0.81 &  0.86 &  0.92 &  0.81 &  0.87 &  0.92 &  0.78 &  0.84 &  0.91 &  0.83 &  0.88 &  0.94 \\ 
Affect &  0.46 &  0.55 &  0.74 &  0.51 &  0.59 &  0.76 &  0.46 &  0.53 &  0.72 &  0.52 &  0.6 &  0.78 \\ 
Activities &  0.31 &  0.38 &  0.79 &  0.34 &  0.4 &  0.81 &  0.29 &  0.36 &  0.77 &  0.36 &  0.41 &  0.82 \\ 
Arousal &  0.51 &  0.56 &  0.53 &  0.52 &  0.56 &  0.56 &  0.51 &  0.55 &  0.54 &  0.53 &  0.57 &  0.57 \\ 
Valence &  0.51 &  0.62 &  0.57 &  0.53 &  0.6 &  0.58 &  0.52 &  0.6 &  0.56 &  0.53 &  0.62 &  0.58 \\ 
Hypertension &  0.69 &  0.73 &  0.74 &  0.64 &  0.69 &  0.69 &  0.7 &  0.73 &  0.74 &  0.71 &  0.75 &  0.77 \\ 
\hline 
\rowcolor{white} 
Avg & 0.55  & 0.62  & 0.71  & 0.56  & 0.62  & 0.72  & 0.54  & 0.6  & 0.71  & \textbf{0.58}  & \textbf{0.64}  & \textbf{0.74}  \\ 
\hline 
Reg ($\downarrow$) & \textbf{MAE} & \textbf{MSE} & \textbf{MAPE} & \textbf{MAE} & \textbf{MSE} & \textbf{MAPE} & \textbf{MAE} & \textbf{MSE} & \textbf{MAPE} & \textbf{MAE} & \textbf{MSE} & \textbf{MAPE} \\ 
\hline 
HR (Dalia) &  10.5 &  221 &  0.12 &  9.47 &  181 &  0.11 &  9.27 &  189 &  0.1 &  7.78 &  143 &  0.09 \\ 
Avg-HR &  4.43 &  34.4 &  0.06 &  4.75 &  37.6 &  0.07 &  3.82 &  26.6 &  0.05 &  3.8 &  26.3 &  0.05 \\ 
Sys-BP &  14.0 &  336 &  0.11 &  14.5 &  347 &  0.12 &  13.6 &  315 &  0.11 &  13.2 &  281 &  0.11 \\ 
Dia-BP &  8.68 &  121 &  0.12 &  8.55 &  118 &  0.12 &  8.53 &  117 &  0.12 &  8.16 &  109 &  0.12 \\ 
Sys-BP (VV) &  15.7 &  439 &  0.12 &  15.5 &  432 &  0.12 &  15.6 &  446 &  0.12 &  15.9 &  451 &  0.12 \\ 
Dia-BP (VV) &  8.12 &  119 &  0.1 &  8.6 &  133 &  0.11 &  8.25 &  127 &  0.1 &  8.04 &  123 &  0.1 \\ 
HR-Green &  9.68 &  188 &  0.13 &  9.5 &  192 &  0.13 &  9.11 &  181 &  0.12 &  7.61 &  149 &  0.1 \\ 
HR-Infrared &  11.6 &  246 &  0.16 &  11.0 &  228 &  0.15 &  11.4 &  248 &  0.16 &  9.82 &  206 &  0.14 \\ 
HR-Red &  12.1 &  261 &  0.17 &  11.8 &  250 &  0.16 &  12.0 &  266 &  0.16 &  10.8 &  231 &  0.15 \\ 
\hline 
\rowcolor{white} 
Avg & 10.5  & 218  & 0.12  & 10.4  & 213  & 0.12  & 10.2  & 213  & 0.12  & \textbf{9.45}  & \textbf{191}  & \textbf{0.11}  \\ 
\hline 
\end{tabular} 
\label{tab:linear-probing-foundation} 
\end{table}

\newpage
\section{Noise robustness evaluation}
\label{app:noise}
\begin{figure}[h]
    \centering
    \includegraphics[width=0.8\linewidth]{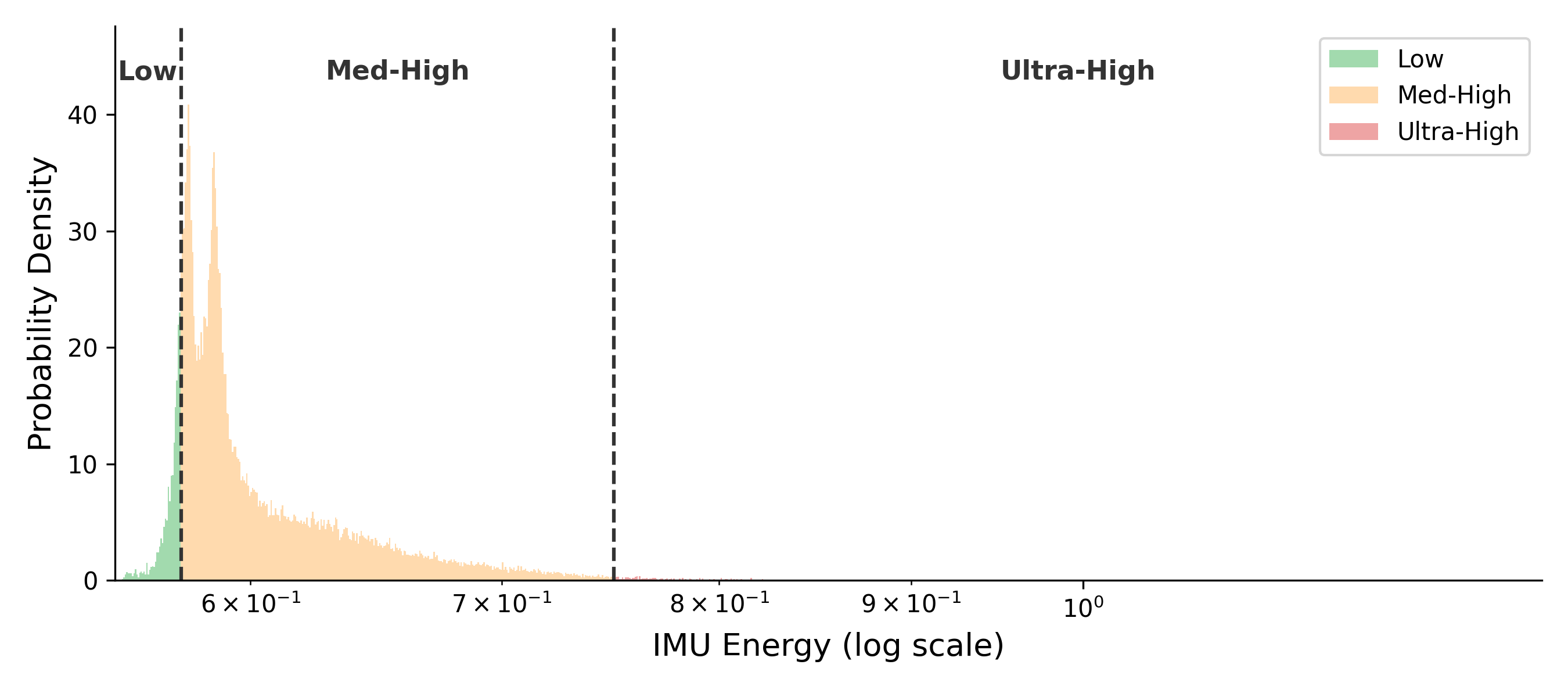}
    \caption{
        The distribution of IMU energy in the DaLiA dataset,
        with the x-axis on a log-scale.
    }
    \label{fig:noise}
\end{figure}

To further evaluate robustness under varying motion conditions, we conduct a noise sensitivity analysis on the PPG-DaLiA dataset using IMU-based motion estimates. We first calculate the IMU energy by taking the square of IMU values along each IMU direction in the 8 second window that corresponds to the PPG data, averaging the directions and timesteps within the window, and then taking the square root. This gives us an IMU energy estimate in each PPG window. The regime definitions (low, med-high, ultra-high) are defined based on the IMU energy distribution shown in Figure~\ref{fig:noise}. To define low noise IMU energy ($< 0.575$), we use any IMU energy that is less common (before the clear peak in IMU energy values in Figure ~\ref{fig:noise}), middle-to-high ($0.575-0.75$) is defined to capture the bulk of the IMU energy distribution, and ultra-high IMU energy values ($> 0.75$) are ones that fall into the long tail of the distribution. These thresholds roughly correspond to distinct operating conditions ranging from relatively stable acquisition to severe motion artifacts. Heart rate regression results in each of these regimes are reported in Table~\ref{tab:noise}. The medium-to-high noise regime, which most closely reflects the most common IMU energy estimate and thus realistic wearable usage conditions, yields the largest relative gains for our method compared to all baselines. This indicates that physiologically grounded supervision is most effective under moderate-to-strong motion contamination, where traditional methods begin to degrade but signals remain partially recoverable. In the ultra-high noise regime, all methods exhibit increased error and variance, consistent with extreme signal corruption. It is notable however, that our approach remains competitive and does not exhibit disproportionate degradation. In the low-noise regime, baseline methods are potentially more sensitive to minor waveform perturbations, whereas our method yields more stable predictions by reducing sensitivity to instantaneous signal fluctuations due to the physiological supervision. Overall, these findings reinforce the robustness claims of our approach and are consistent with the main experimental results reported in the paper.

\begin{table}[h]
    \centering
    \caption{
        Model performance by noise level
        (MAE $\pm$ sample count).
    }
    \label{tab:noise}

    \begin{tabular}{lrrrrrr}
        \toprule
        \textbf{Model} &
        \multicolumn{2}{c}{Low} &
        \multicolumn{2}{c}{Med-High} &
        \multicolumn{2}{c}{Ultra} \\
        
        & \textbf{MAE} & \textbf{N}
        & \textbf{MAE} & \textbf{N}
        & \textbf{MAE} & \textbf{N} \\
        
        \midrule
        PaPaGei-S & 9.35 & 6K & 15.65 & 58K & 19.80 & 789 \\
        PaPaGei-P & 7.91 & 6K & 13.17 & 58K & 16.42 & 789 \\
        PulsePPG  & 5.82 & 6K & 10.96 & 58K & 13.17 & 789 \\
        \textbf{Ours} & 4.13 & 6K & 8.09 & 58K & 12.01 & 789 \\
        \bottomrule
    \end{tabular}
\end{table}

\end{document}